\documentclass[sigconf]{acmart}

\usepackage[utf8]{inputenc} 
\usepackage[T1]{fontenc}    
\usepackage{hyperref}       
\usepackage{url}            
\usepackage{booktabs}       
\usepackage{amsfonts}       
\usepackage{nicefrac}       
\usepackage{microtype}      
\usepackage{amsmath}

\usepackage{multicol}
\usepackage{multirow}
\usepackage{makecell}
\usepackage{arydshln}
\usepackage{array}
\usepackage{bm}
\usepackage{algorithm}
\usepackage{algorithmic}
\usepackage{amsmath}
\usepackage{graphicx}
\usepackage{subfig}
\usepackage[labelfont=bf]{caption}
\usepackage{sidecap}
\usepackage{dblfloatfix}

\usepackage{pifont}  
\newcommand{\cmark}{\ding{52}}  
\newcommand{\xmark}{\ding{53}}  
\newcommand{\correct}{\textcolor{green!60!black}{\cmark}}
\newcommand{\incorrect}{\textcolor{red!80!black}{\xmark}}

\newcommand{\AXE}{AXE }
\newcommand{\axe}{AXE}
\newcommand{\hidek}[1]{}

\copyrightyear{2025}
\acmYear{2025}
\setcopyright{cc}
\setcctype{by}
\acmConference[FAccT '25]{The 2025 ACM Conference on Fairness,
Accountability, and Transparency}{June 23--26, 2025}{Athens, Greece}
\acmBooktitle{The 2025 ACM Conference on Fairness, Accountability, and
Transparency (FAccT '25), June 23--26, 2025, Athens,
Greece}
\acmDOI{10.1145/3715275.3732219}
\acmISBN{979-8-4007-1482-5/2025/06}

\begin{document}


\title{Evaluating Model Explanations without Ground Truth}

\author{Kaivalya Rawal}
\email{kaivalyarawal45@gmail.com}
\orcid{0000-0003-1087-4599}
\affiliation{
  \institution{Oxford Internet Institute}
  \city{Oxford}
  \state{Oxfordshire}
  \country{UK}
}

\author{Zihao Fu}
\email{zihao.fu@oii.ox.ac.uk}
\affiliation{
  \institution{Oxford Internet Institute}
  \city{Oxford}
  \state{Oxfordshire}
  \country{UK}
}

\author{Eoin Delaney}
\email{eoin.delaney@tcd.ie}
\affiliation{
  \institution{Trinity College Dublin}
  \city{Dublin}
  \state{}
  \country{Ireland}
}

\author{Chris Russell}
\email{chris.russell@oii.ox.ac.uk}
\affiliation{
  \institution{Oxford Internet Institute}
  \city{Oxford}
  \state{Oxfordshire}
  \country{UK}
}



\begin{abstract}

    There can be many competing and contradictory explanations for a single model prediction, making it difficult to select which one to use. Current explanation evaluation frameworks measure quality by comparing against ideal ``ground-truth'' explanations, or by verifying model sensitivity to important inputs. We outline the limitations of these approaches, and propose three desirable principles to ground the future development of explanation evaluation strategies for local feature importance explanations. We propose a ground-truth \textbf{Agnostic eXplanation Evaluation} framework (\textbf{\axe}) for evaluating and comparing model explanations that satisfies these principles. Unlike prior approaches, \AXE does not require access to ideal ground-truth explanations for comparison, or rely on model sensitivity -- providing an independent measure of explanation quality. We verify \AXE by comparing with baselines, and show how it can be used to detect explanation fairwashing. Our code is available at \url{https://github.com/KaiRawal/Evaluating-Model-Explanations-without-Ground-Truth}.

\end{abstract}





\maketitle

\section{Introduction}
\label{sec:introduction}

\begin{figure*}[ht]
    \centering
    \vspace{-0.2cm}
    \subfloat[\textbf{``Gradients''} explainer: \textbf{Diabetes Pedigree Function} is the most important input feature, pushing the classifier to a positive prediction (diabetic)]{
        \includegraphics[width=0.48\linewidth]{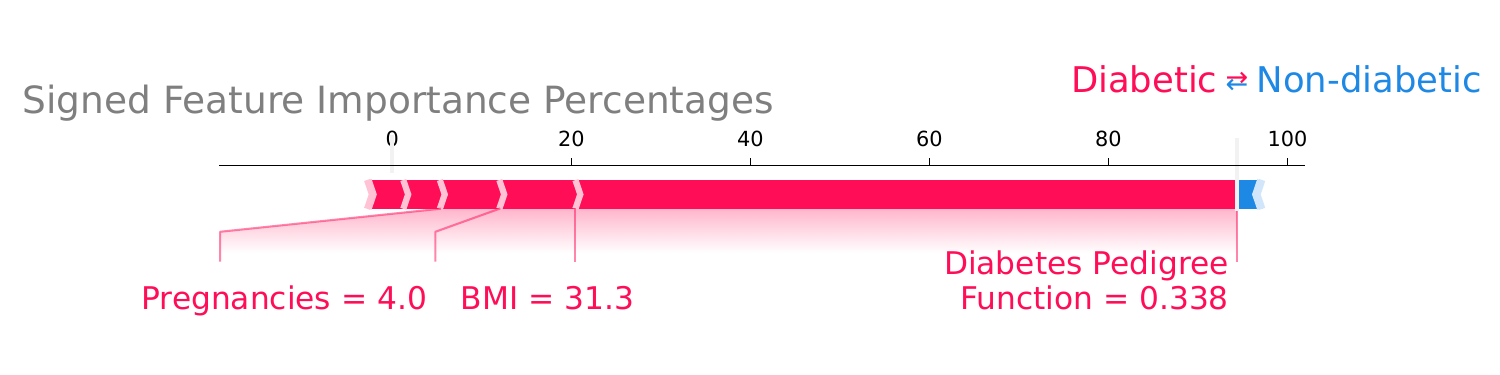}
    }
    \hfill
    \subfloat[\textbf{``SHAP''} explainer: \textbf{Glucose} is still the most important (positive) input, \\ \textbf{BMI} now has an negative importance (non-diabetic)]{
        \includegraphics[width=0.48\linewidth]{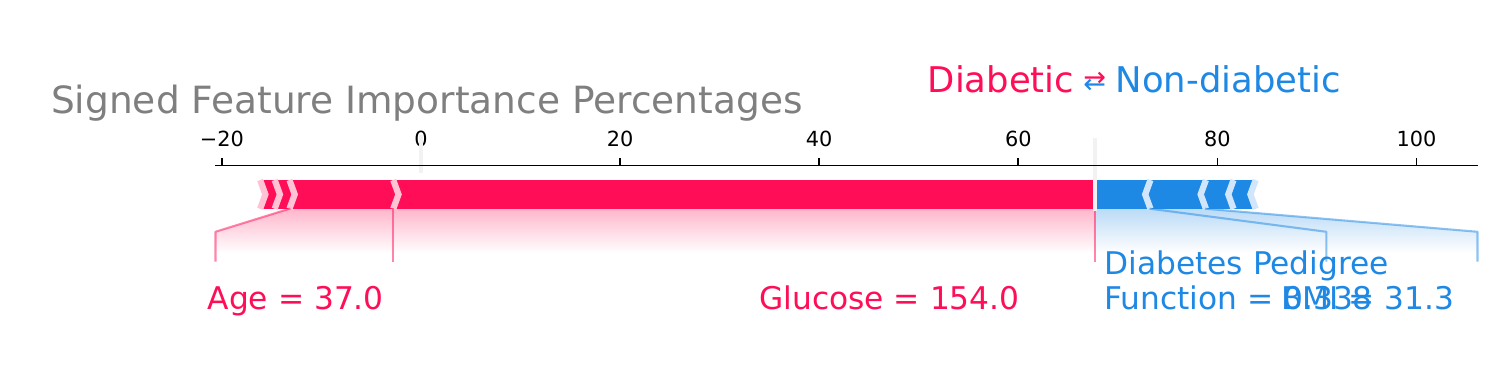}
    } \\
    \hfill
    \vspace{-0.2cm}
    \subfloat[\textbf{``LIME''} explainer: \textbf{Glucose} (positive), \textbf{BMI} (positive), and \textbf{Insulin} \\ (negative) are all important input features]{
        \includegraphics[width=0.48\linewidth]{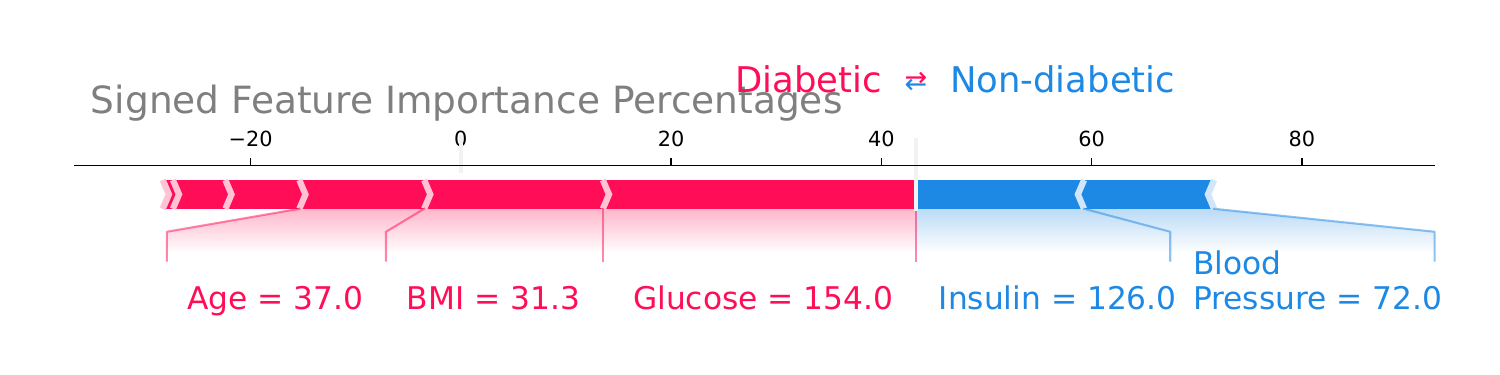}
    }
    \hfill
    \subfloat[\textbf{``Integrated Gradients''} explainer: \textbf{Glucose} (positive) \textbf{BMI} (positive) and \textbf{Blood Pressure} (negative) are important.]{
        \includegraphics[width=0.48\linewidth]{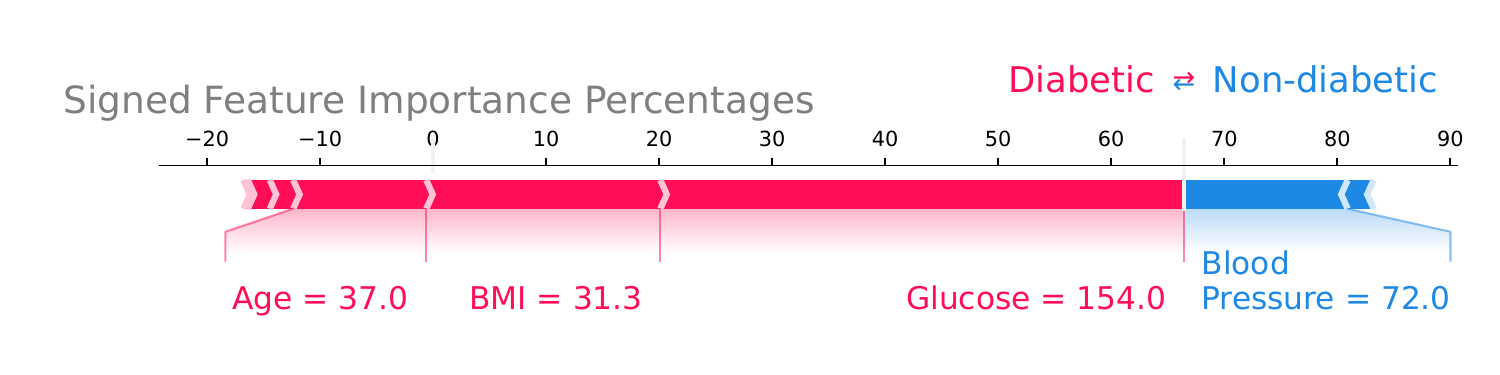}
    }
     \caption{\textbf{Different Explainers Yield Different Explanations}: A neural network predicts diabetes on the ``Pima Indians'' dataset \cite{pima}. A single positive (diabetic) prediction is explained using four explainers. These feature-importance explanations, visualized here as ``force-plots'', consist of a signed vector indicating the relative contribution of each input to the model output. They disagree with each other. Section \ref{subsec:prior} details the explainers, and section \ref{subsec:framework} evaluates these four explanations using \axe.}
     \Description[]{}
    \label{fig:disagreement_example}
\end{figure*}

As artificial intelligence (AI) systems are increasingly used in critical decision-making processes, knowing which model explanation to trust has emerged as a fundamental challenge. Model explanations often disagree with each other (see figure \ref{fig:disagreement_example}), and the selection of incorrect or intentionally misleading explanations can have far-reaching consequences -- from misinforming users and regulators to reinforcing systemic biases and eroding public trust in AI systems \cite{disagreement_original, disagreement_new, rashomon}. This challenge is particularly acute in high-stakes domains like healthcare diagnostics, financial and credit scoring services, and criminal justice, where machine learning models directly impact human lives. It is essential to be able to select the best explanation from a set of possible explanations, but unfortunately there has been little progress towards this critical problem.

User studies offer a workaround -- of approximately 300 papers proposing new model explanation methods (explainers), one in seven performed user study evaluations \cite{state_of_xai_evals}. However, one in three papers evaluated entirely anecdotally, reflecting the need for standardized evaluation frameworks \cite{state_of_xai_evals}. Without consensus on the essential properties that explanations should possess and robust frameworks to numerically evaluate them, progress in the field remains slow and fragmented. Historically, advances in AI have often been driven by benchmark datasets and deterministic evaluation frameworks defined using standard metrics -- as exemplified by ImageNet for computer vision \cite{imagenet} and MMLU for language understanding \cite{mmlu}. Developing analogous benchmarks for eXplainable AI (XAI) involves unique challenges, including a lack of access to reliable ``ground-truth'' explanations to compare against. This hampers our ability to meaningfully evaluate competing explanation methods, assess their utility to users impacted by AI systems, or their faithfulness to model behavior. Progress towards these goals can ensure that XAI truly makes AI systems transparent.

There are many forms of model explanation. One popular category among practitioners is post-hoc model-agnostic feature-importance explanations, such as LIME \cite{lime} or SHAP \cite{shap}. These provide explanations for individual predictions rather than describing global model behavior. They can operate on any model type, including neural networks, regardless of weights or architecture. They produce feature importances as output: a signed vector indicating the relative contribution of each input feature to the output. While there are many competing explanation types, data modalities, and evaluation desiderata, \cite{DBLP:journals/corr/abs-2110-10790}, this paper focuses exclusively on local feature-importance explanations for models operating on tabular datasets. Even in this restricted setting, different explanation methods (explainers) often provide contradictory explanations (figure \ref{fig:disagreement_example}). In this paper we do not propose a new XAI method but instead develop three general principles: \emph{local contextualization}, \emph{model relativism}, and \emph{on-manifold evaluation} to guide the evaluation of feature-importance explanations. We use these to propose \axe, a new ground-truth \textbf{Agnostic eXplanation Evaluation} framework that considers a good explanation to be one that correctly identifies the features most predictive of model outputs. \AXE is inspired by user research which indicates useful explanations are those that help users emulate and predict model behavior \cite{predictions_for_explanations}. 

The plots in figure \ref{fig:disagreement_example} visualize competing explanations for the same datapoint in a diabetes classification model. They disagree with each other in the contributions of the input features, a phenomenon commonly documented in the literature \cite{disagreement_original, disagreement, disagreement_new}. Some XAI methods such as LIME and SHAP often rely on off-manifold model predictions to generate a single explanation, leading to different explanations. This explicit reliance on feature sensitivity is one potential cause for explanation disagreement, which we seek to address through the \emph{on-manifold evaluation} principle proposed in section \ref{subsec:principles} A good explanation evaluation framework should provide clear guidance about which explanation is better, helping users make sense of competing explainers. Explanation disagreement can be exploited by adversaries to produce fairwashed explanations -- where a given explainer certifies that protected attributes were not important to the model even if they determined the prediction \cite{fairwashing, adv_attack}. This presents a risk for auditors and regulators enforcing AI fairness, further motivating our work and highlighting the importance of evaluating explanation quality.

This paper is structured as follows: in section \ref{sec:prelims} we define our notation, introduce three foundational principles for explanation evaluation, and describe prior work. In section \ref{sec:methodology} we introduce \axe, an explanation evaluation framework directly couched in terms of predictive accuracy -- the notion that a good human-interpretable explanation is one which identifies the features most predictive of the model behavior \cite{xai_predictiveness, predictions_for_explanations}. In section \ref{sec:experiments}, we demonstrate how \AXE can be used to detect explanation fairwashing -- foiling a state-of-the-art adversarial attack \cite{adv_attack}, and compare \AXE with existing baselines from the literature. We conclude in section \ref{sec:conclusion} with a brief summary discussion.

\section{Evaluating Model Explanations}
\label{sec:prelims}

\begin{table*}[b!]
\centering
\caption{\textbf{Explanation Evaluation Metrics}: Definitions for ground-truth based explanation evaluation metrics: FA, RA, SA, SRA, RC and PRA \cite{disagreement, agarwal2022openxai} ($\mathbf{e}$ is an explanation, and $\mathbf{e}^{*}$ is the ground truth); sensitivity based metrics PGI and PGU \cite{agarwal2022openxai, pgi_intro, pgi_counterfactuals}; and \axe. For each we list whether it satisfies the three evaluation principles laid out in section \ref{subsec:principles}. \textit{* For PGU, lower values are better}}
\vspace{-0.2cm}
\begin{tabular}{lm{6.7cm}ccc}
\toprule
\textbf{Metric} & \textbf{Definition} & \thead{\textbf{Local}\\\textbf{Contextualization}} & \thead{\textbf{Model}\\\textbf{Relativism}} & \thead{\textbf{On-Manifold}\\\textbf{Evaluation}} \\
\midrule
\makecell[l]{\textbf{FA:} Feature Agreement} & Fraction of top-n features common between $\mathbf{e}$ and $\mathbf{e}^{*}$. & \incorrect & \incorrect & \correct \\
\hdashline[1pt/2pt]
\makecell[l]{\textbf{RA:} Rank Agreement} & Fraction of top-n features common between $\mathbf{e}$ and $\mathbf{e}^{*}$ with the same position in respective rank orders. & \incorrect & \incorrect & \correct \\
\hdashline[1pt/2pt]
\makecell[l]{\textbf{SA:} Sign Agreement} & Fraction of top-n features common between $\mathbf{e}$ and $\mathbf{e}^{*}$ with the same sign. & \incorrect & \incorrect & \correct \\
\hdashline[1pt/2pt]
\makecell[l]{\textbf{SRA:} Signed Rank Agreement} & Fraction of top-n features common between $\mathbf{e}$ and $\mathbf{e}^{*}$ with the same sign and rank. & \incorrect & \incorrect & \correct \\
\hdashline[1pt/2pt]
\makecell[l]{\textbf{RC:} Rank Correlation} & Spearman's rank correlation coefficient for feature rankings from $\mathbf{e}$ and $\mathbf{e}^{*}$. & \incorrect & \incorrect & \correct \\
\hdashline[1pt/2pt]
\makecell[l]{\textbf{PRA:} Pairwise Rank Agreement} & Fraction of feature pairs for which relative ordering in $\mathbf{e}$ and $\mathbf{e}^{*}$ is the same. & \incorrect & \incorrect & \correct \\
\hdashline[1pt/2pt]
\makecell[l]{\textbf{PGI:} Prediction-Gap on Important\\Feature Perturbation} & Mean absolute change in model output upon perturbing top-n most important inputs. & \correct & \correct & \incorrect \\
\hdashline[1pt/2pt]
\makecell[l]{\textbf{PGU*:} Prediction-Gap on\\Unimportant Feature Perturbation} & Mean absolute change in model output upon perturbing top-n most unimportant inputs. & \correct & \correct & \incorrect \\
\hdashline[1pt/2pt] 
\makecell[l]{\textbf{\axe:} (ground-truth) Agnostic\\ eXplanation Evaluation} & Predictiveness of the top-n most important inputs in recovering model output. \textbf{Defined in section \ref{subsec:framework}.} & \correct & \correct & \correct \\
\bottomrule
\end{tabular}
\label{tab:prior_metrics}
\vspace{-0.3cm}
\end{table*}

\begin{figure}[h!]
    \centering
    \vspace{-0.25cm}
    \subfloat[\textbf{Explanation generation}: Explainer $\mathcal{E}$ produces an explanation vector $\mathbf{e}$ of signed feature importances using datapoint $\mathbf{x}$, model $m$ and prediction $m(\mathbf{x})$]{
        \includegraphics[width=0.98\linewidth]{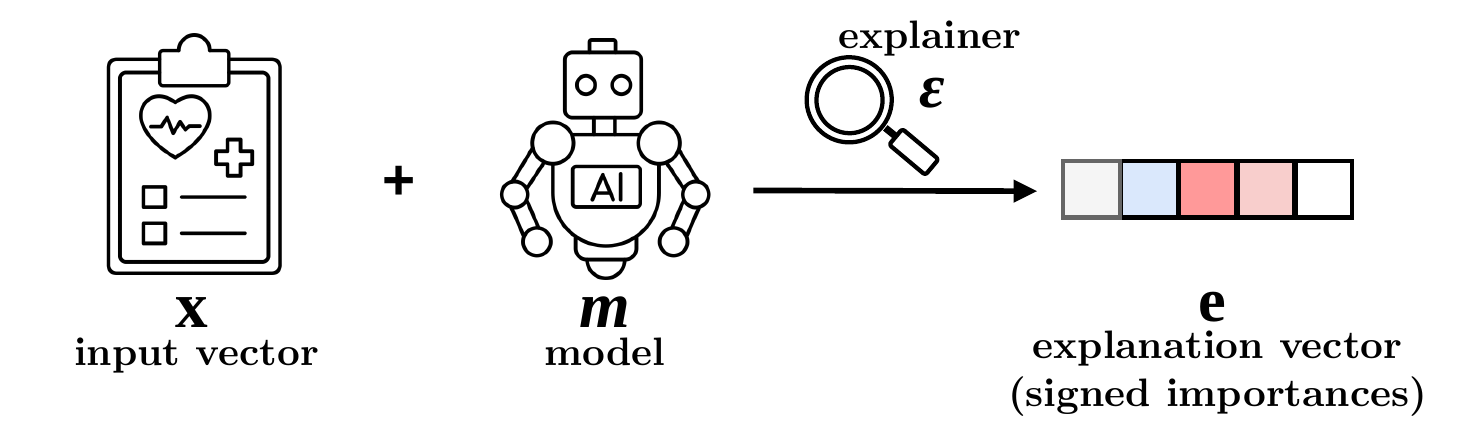}
    }
    \hfill
    \vspace{-0.1cm}
    \subfloat[\textbf{Explanation evaluation}: \AXE evaluates the quality $q$ of explanation $\mathbf{e}$ by measuring how accurately prediction $m(\mathbf{x})$ can be recovered from dataset $\mathcal{X}$.]{
        \includegraphics[width=0.98\linewidth]{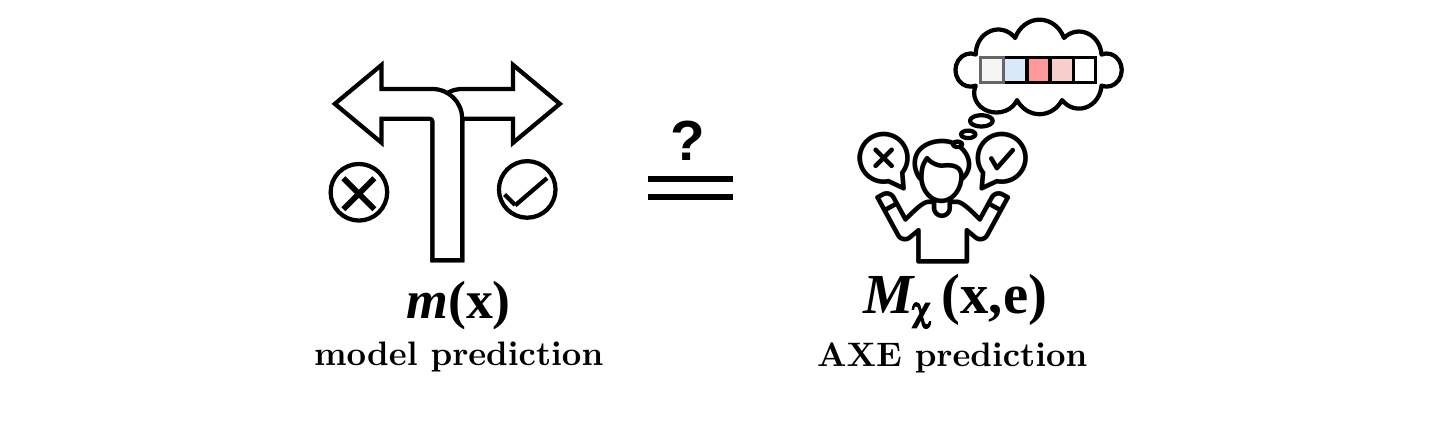}
    }
    \vspace{-0.2cm}
    \caption{\textbf{Explanation Generation (a) and Evaluation (b)}: \AXE measures how well a given explanation can help emulate model behavior. See section \ref{subsec:framework} for full algorithm. }
    \Description[]{}
    \label{fig:infographic}
    \vspace{-0.2cm}
\end{figure}

A typical scenario depicting the generation and evaluation of local, post-hoc, model-agnostic explanations is presented in figure \ref{fig:infographic}.

\subsection{Our Notation}
\label{subsec:notation}

We specify our notation from figure \ref{fig:infographic}: input vector $\mathbf{x}$, model $m$, and explanation $\mathbf{e}$. We use these to define an explanation ``quality'' metric $q$ and an evaluation framework $Q$ here, and in algorithm \ref{alg:axe} we implement such a framework using \axe.

\begin{enumerate}
    \item \emph{Input Vector}: The input feature vector for any arbitrary datapoint is defined as:
\(
\mathbf{x} = [x_1, x_2, \dots, x_N] \in \mathbb{R}^N
\),
where \( N \) is the number of features, and each \( x_i \) is a real-valued feature.
    \item \emph{Model and Prediction}: The model \( m \) is a mapping from the feature space to a binary output:
\(
m: \mathbb{R}^N \to \{0, 1\}
\),
and the prediction for input \( \mathbf{x} \) is given by:
\(
m(\mathbf{x}) = y_\text{pred} \in \{0, 1\}.
\)
    \item \emph{Explanation}: A local feature importance explanation is denoted \( \mathbf{e} \). It is a function of the input \( \mathbf{x} \) and model \( m \) (implicitly model prediction \( m(\mathbf{x}) \) too). For an explainer $\mathcal{E}$,
\(
\mathbf{e} = \mathcal{E}(\mathbf{x}, m)
\), 
where \( \mathbf{e} \in \mathbb{R}^N \), and each component \( e_i \) represents the (signed) contribution or importance of the feature \( x_i \) to the prediction \( m(\mathbf{x}) \).
    \item \emph{Explanation Quality Metric}: For dataset $\mathcal{X}$, the explanation quality metric \( q \in [0,1] \) evaluates the quality of explanation \( \mathbf{e} \) for a specific input \( \mathbf{x} \) and model \( m \). As a function,  $q=q_\mathcal{X}(\mathbf{x},m,\mathbf{e})$ where $0 \leq q \leq 1$ (greater $q$ is better). Previous work often refers to quality scores as fidelity or explanation faithfulness \cite{electronics8080832, abacus_xai_bench, gilpin_xai_review}.
\end{enumerate}

An \emph{explanation evaluation framework} is a tuple \( (\mathcal{X}, m, \mathcal{E}, Q) \):
\begin{itemize}
    \item \( \mathcal{X} \in \mathbb{R}^{\nu \times N}\) is the dataset of inputs with $N$ features and $\nu$ datapoints, \( \mathcal{X} = \{\mathbf{x}_1, \mathbf{x}_2, \dots, \mathbf{x}_\nu \} \).
    \item \( m : \mathbb{R}^N \to \{0, 1\}\) is the model being explained.
    \item \( \mathcal{E} : \mathbb{R}^N \to \mathbb{R}^N \) is the explanation method that generates explanation \( \mathbf{e} \in \mathbb{R}^N \) for each datapoint \( \mathbf{x} \in \mathbb{R}^N \).
    \item \( Q \) is the aggregate quality score over the dataset computed as an average of explanation quality $q$:
\end{itemize}
\vspace{-0.3cm}
    \[
    Q(\mathcal{X}, m, \mathcal{E}) = \frac{1}{\nu} \sum_{i=1}^\nu q(\mathbf{x}_i, m, \mathcal{E}(\mathbf{x}_i, m),\mathcal X).
    \]

\subsection{Three Principles for Evaluating Model Explanation Quality}
\label{subsec:principles}

Variations in explanations occur for many reasons. For example: (a) different input datapoints $\mathbf{x_1} \neq \mathbf{x_2}$ typically have different explanations; (b) different prediction models $m_1 \neq m_2$ -- eg. with updated neural network weights -- typically have different explanations; and (c) as seen in figure  \ref{fig:disagreement_example}, explanations from different explainers can have different explanations, possibly due to varying off-manifold input sensitivity of the model $m$ in the neighborhood of $\mathbf{x}$. An evaluation framework that cannot distinguish between explanations from these varying scenarios and always scores diverse explanations the same is not helpful. We characterize these situations respectively with the following principles:

\begin{figure*}[hb!]
\vspace{-0.1cm}
    \centering
    \subfloat[Rank Agreement: $RA_{n=2}$]{
        \includegraphics[width=0.32\linewidth]{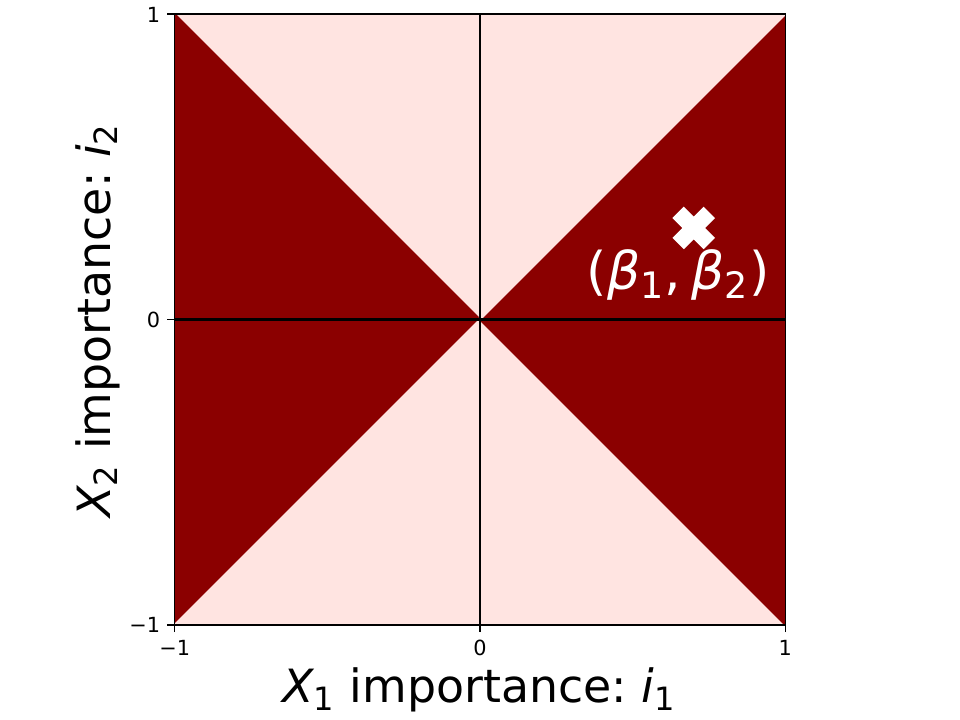}
    }
    \hfill
    \subfloat[Sign Agreement: $SA_{n=2}$]{
        \includegraphics[width=0.32\linewidth]{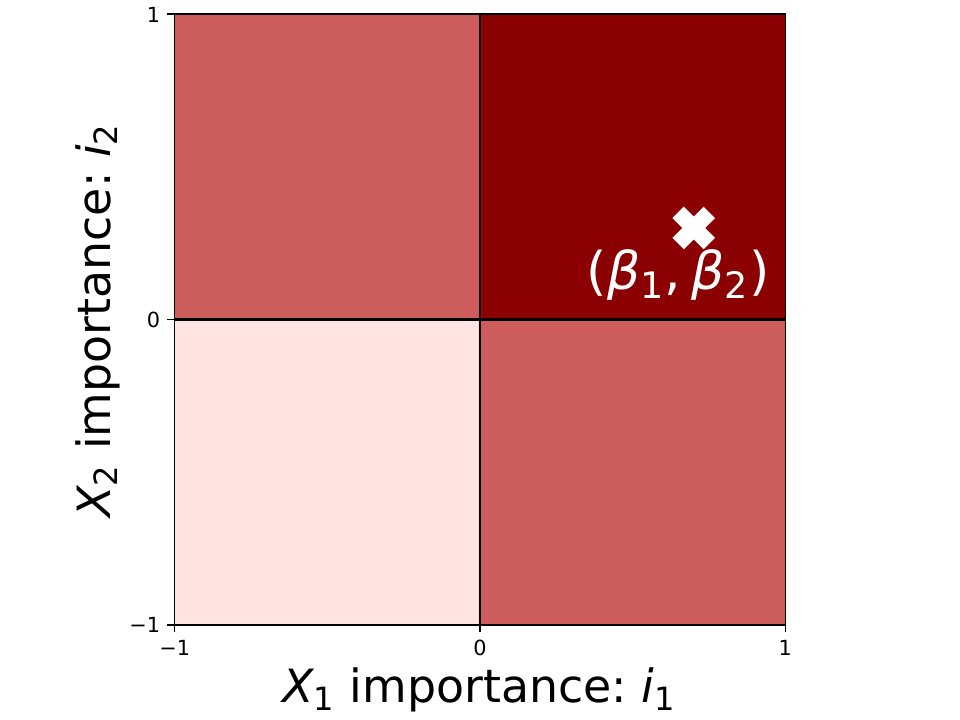}
    }
    \hfill
    \subfloat[Signed Rank Agreement: $SRA_{n=2}$]{
        \includegraphics[width=0.32\linewidth]{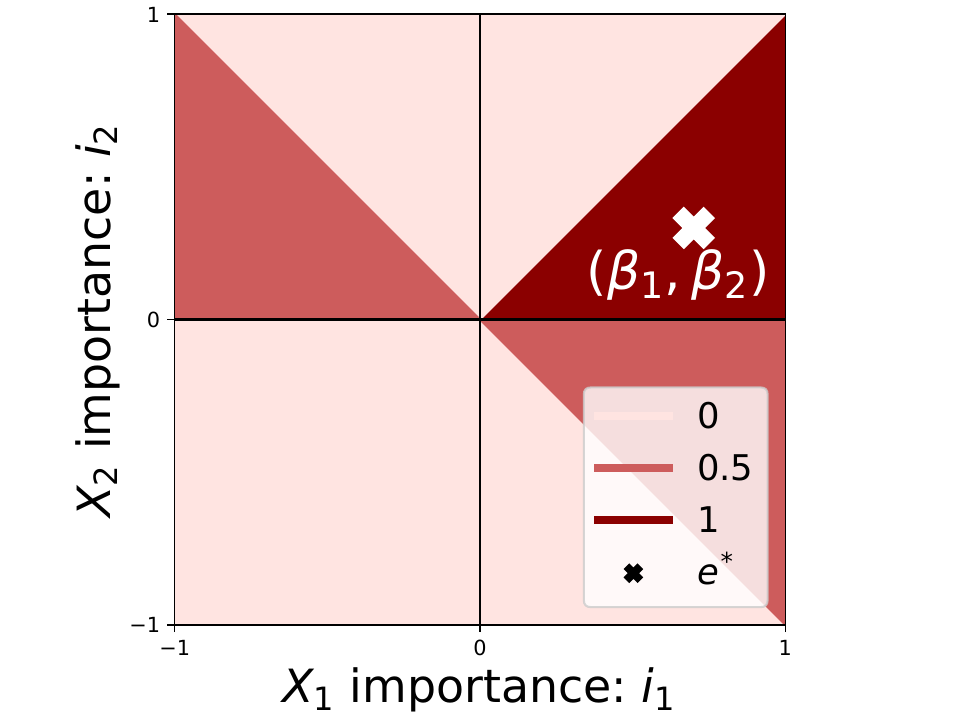}
    }
    \caption{\textbf{Violations of local contextualization and model relativism}: Plots showing explanation quality $q$ (color) across $i_1$ and $i_2$ values for explanation $\mathbf{e} = (i_1, i_2)$. Model $m(\mathbf{x})=\beta_0 + \beta_1 X_1 + \beta_2 X_2$ has ground-truth $\mathbf{e}^{*} = (\beta_1, \beta_2) = (0.7,0.3)$. Diverse explanations $\mathbf{e}$ map to the same quality $q$ (0, 0.5, or 1), violating \emph{local contextualization}. Changing the model changes the ground-truth $\mathbf{e}^{*}$, but leaves the plots unchanged $\forall \beta_1, \beta_2$ where $\beta_1 > \beta_2 > 0$, violating \emph{model relativism}. Section 
    \Description[]{}
    \ref{subsec:gt_failures} explains these computations.}
    \label{fig:gt_failures}
    \vspace{-0.3cm}
\end{figure*}

\begin{enumerate}
    \item \textbf{Local Contextualization}: \emph{Explanations should depend on the datapoint being explained.} For local explanations, when the datapoint $\mathbf{x}$ changes, the evaluation metric $q$ should not always prefer that the corresponding explanation $\mathbf{e}$ remain unchanged. Model behavior is not always identical across the data distribution.
    \item \textbf{Model Relativism}: \emph{Explanations should depend on the model being explained.} When the model $m$ changes, the evaluation metric $q$ should not always prefer that the corresponding explanation $\mathbf{e}$ remain unchanged.
    \item \textbf{On-manifold Evaluation}: \emph{Explanations on-manifold should not depend on changes in off-manifold model behavior.} When off-manifold model predictions $m(\mathbf{x}+\delta\mathbf{x})$ change, the evaluation metric $q$ should remain unchanged for explanation $\mathbf{e}$. Evaluation metrics should not make the same assumptions as the explainers they seek to evaluate -- which often assume changes in output caused by synthetic perturbations in particular model inputs indicate the importance of those inputs. 
\end{enumerate}

The \emph{on-manifold evaluation} principle is motivated by the observation that many explanation methods are variants of sensitivity analysis that capture how much synthetically varying a particular feature alters model outputs \cite{sens_in_xai, ood_sens_in_xai, efficient_sens_in_xai}. Ideally, an explanation for model behavior on datapoint $\mathbf{x_1}$ should not depend on model behavior on a different datapoint $\mathbf{x_2}=\mathbf{x}+\delta\mathbf{x}$. Further, evaluation frameworks that capture the fidelity of explanations with respect to synthetic neighborhoods around real points, are simply encoding a particular choice of sensitivity analysis without meaningfully evaluating the explanation quality. Section \ref{subsec:sens_failures} formalizes this.

Previous methods for computing the quality $q$ of explanation $\mathbf{e}$ have suggested comparing $\mathbf{e}$ with a known ``ground-truth'' vector $\mathbf{e}^{*}$. Proposals include one ``ground-truth'' per datapoint $\mathbf{x}$, unintentionally introducing independence from $m$ \cite{pointinggame, salsanity}; or one ``ground-truth'' per model $m$, introducing independence from $\mathbf{x}$ \cite{disagreement, agarwal2022openxai}. The latter case clearly violates \emph{local contextualization} by comparing each local explanation with the same static ``ground-truth'', promoting a holistic global model explanation instead of local explanations that differ across datapoints. The former case violates \emph{model relativism} by computing quality $q$ for explanation $\mathbf{e}$ using an immutable ``ground-truth'' $\mathbf{e}^{*}$, fixed for a given datapoint, regardless of the model used. With images especially, an explanation is often considered good if it selects the ``correct'' region as important in an image -- regardless of whether the model used those features \cite{pointinggame, salsanity, psycho_pointing_evaluation}. Section \ref{subsec:gt_failures} showcases these violations in detail.

\subsection{Prior Approaches}
\label{subsec:prior}

Explanations can be evaluated using any of the evaluation metrics defined in table \ref{tab:prior_metrics}. Broadly, these fall into two categories: 

\begin{enumerate}
    \item \emph{ground-truth} based  metrics compare the generated explanations $\mathbf{e}$ with ground-truth annotations $\mathbf{e}^{*}$, either collected by humans or inferred using a different proxy \cite{rise, pert_xai_1, pert_xai_2, gradcam}. These include Feature Agreement (FA), Sign Agreement (SA), Rank Agreement (RA), Signed Rank Agreement (SRA), Rank Correlation (RC) and Pairwise Rank Agreement (PRA) \cite{disagreement}.
    \item \emph{sensitivity} based metrics verify model sensitivity to the inputs declared important by an explanation \cite{rise, efficient_sens_in_xai, xai_visual_eval, xrai}. These have been summarized as Prediction Gap on Important Feature Perturbation (PGI) and Prediction Gap on Unimportant Feature Perturbation (PGU) \cite{pgi_intro, pgi_counterfactuals, agarwal2022openxai}.
\end{enumerate}

In addition to evaluation metrics $q$, we also summarize the most common explanation methods (explainers) $\mathcal{E}$. We limit ourselves to post-hoc explainers that produce signed feature importance vectors as explanations. Many of these are inspired by sensitivity analysis -- measuring how changes in input variables effect changes in the model response \cite{sens_in_xai, ood_sens_in_xai, efficient_sens_in_xai}. Both LIME and SHAP sample synthetic points in the neighborhood of a given datapoint $\mathbf{x}$ and fit linear models to obtain feature importances \cite{lime, shap}. Gradient-based methods compute the gradient of the model output with respect to the input \cite{grad}, with several extensions: Smooth Grad \cite{smoothgrad}, Integrated Gradients \cite{integratedgradients}, and Input \(\times\) Grad \cite{itg}. In our experiments in section \ref{subsec:openxai_experiments}, we used the standard OpenXAI benchmark for generating explanations $\mathbf{e}$ and evaluating them using prior metrics $q$. \cite{agarwal2022openxai}.

Several user studies have demonstrated that for explanations to be useful in real-world scenarios, their primary function must be to help users predict model behavior \cite{predictions_for_explanations, user_predictiveness_1, user_predictiveness_2, user_predictiveness_3, user_predictiveness_4, user_predictiveness_5}. \AXE is designed to explicitly operationalize this idea, filling a critical gap in the literature. While work has referenced the need to move beyond ground-truth and sensitivity towards predictiveness as a measure of explanation quality \cite{xai_predictiveness, doshivelez2017rigorousscienceinterpretablemachine, lipton_on_xai}, few metrics have implemented this idea. Some previous instantiations have been used to measure the quality of explanations in image classification \cite{axe_in_images, psycho_pointing_evaluation}, where \emph{model relativism} violations are common and egregious \cite{salsanity, pointinggame, psycho_pointing_evaluation}.

\subsection{Invariance of ground-truth Metrics to Changing Data and Models}
\label{subsec:gt_failures}

Real world situations lack access to an oracle to provide ground-truth explanations $\mathbf{e}^{*}$ \cite{electronics10050593, missing_ground_truths}. For linear models, a common resolution adopts the model coefficients as the ``ground-truth'' for all datapoints $\mathbf{x} \in \mathcal{X}$ in a dataset \cite{disagreement, agarwal2022openxai}. As mentioned in section \ref{subsec:principles}, comparing with the same $\mathbf{e}^{*} \forall \mathbf{e} \in E$ is undesirable for local explanations and promotes a single explanation across datapoints.

Figure \ref{fig:gt_failures} depicts such an example, directly violating the \emph{local contextualization} principle. Consider a model with two input features, $X_1$ and $X_2$, and prediction $y$, parameterized $y = \beta_0 + \beta_1 X_1 + \beta_2 X_2$. For datapoint $\mathbf{x}$ feature importances are $i_1$ and $i_2$, with explanation $e = [i_1, i_2]$. FA, RA, SA, SRA, RC, and PRA measure explanation quality by comparing with ground-truth explanation $\mathbf{e}^{*} = [\beta_1, \beta_2]$ \cite{disagreement}. Since $\mathbf{e}^{*}$ is constant and independent of $\mathbf{x}$, every explanation is compared against the same tuple $[\beta_1, \beta_2]$. This comparison takes many forms, with definitions provided in table \ref{tab:prior_metrics}. For example, our $N=2$ feature setup implies that for the top $n$ features: $\text{FA}_{n=0} = 0$, $\text{FA}_{n=1} \in \{0, 0.5, 1.0\}$, and $\text{FA}_{n=2} = 1$, and that $\text{FA}_{n=1} = \text{RA}_{n=2} = \text{PRA}_{n=2}$, while RC is undefined.

Plotting evaluation metric $q$ for all possible explanations $e = [i_1, i_2]$, for an example model with $\beta_1 = 0.7$ and $\beta_2 = 0.3$, we see that regardless of the specific value of $\mathbf{e}$, there are regions where the resulting $RA_{n=2}$ is the same (figure \ref{fig:gt_failures} a). Similarly, the $SA_{n=2}$ is the same across $i_1, i_2$ regions (figure \ref{fig:gt_failures} b) and $SRA_{n=2}$ too (figure \ref{fig:gt_failures} c). Concretely, any explanation $e = (i_1, i_2)$ such that $i_1 > 0$, $i_2 > 0$, and $i_1 > i_2$ (this is the region labeled 1 in figure \ref{fig:gt_failures} c) is guaranteed to have the same FA, RA, SA, SRA, RC, and PRA. These plots are specific to our particular model and ground-truth, but display multiple regions of constant FA, RA, SA, SRA, and PRA values, displaying a \emph{violation of the local contextualization principle}. In real world settings, these metrics would fail to distinguish different explanations from each other in quality if they belonged to the same region in figure \ref{fig:gt_failures}.

The same logic demonstrates violations of \emph{model relativism}. From table \ref{tab:prior_metrics} we can see that all ground-truth comparison metrics are symmetric. The metrics are invariant to changes in $\mathbf{e}^{*}$, the same way they are invariant to changes in $\mathbf{e}$. For a given explanation $\mathbf{e}$, and for any model $m$ such that $\beta_1 > \beta_2 > 0$, the plots in figure \ref{fig:gt_failures} would stay unchanged. Concretely, while we used model weights (ignoring $\beta_0$) $m = (\beta_1, \beta_2) = (0.7, 0.3)$, the plots would be unchanged for $m_a = (0.99, 0.01)$, $m_b =(0.99, 0.98)$, and $m_c = (0.02, 0.01)$ -- vastly different models! The importance of $X_1$ with respect to $X_2$ ranges from $1$ to $\infty$ in the limit, and it is absurd for an explanation quality metric $q$ to be unchanged for these diverse models. In this way, FA, RA, SA, SRA, RC, and PRA \emph{fail the model relativism principle}, no longer distinguishing explanations by quality when the underlying model changes.

\vspace{-0.05cm}
\subsection{Inherent biases in sensitivity Metrics}
\label{subsec:sens_failures}
Sensitivity based explainers $\mathcal{E}$ like LIME are highly sensitive to hyperparameters. This facilitates adversarial fairwashing attacks (section \ref{subsec:adv_attack}) \cite{adv_attack} and can cause feature importances to switch arbitrarily from highly positive to highly negative \cite{molnar_book, Riley2023-rs}. Sensitivity analysis based evaluation metrics $q$ suffer similar problems.

Metrics like PGI and PGU may simply encode a preference for particular explainers. We formalize this in the context of synthetic data. Consider a wide range of explainability measures that measure some loss, $\ell$, defined in terms of the fidelity $F$ to classifier responses \(m(\cdot)\), in a synthetic neighborhood ${\mathcal N}_{\mathbf{x}}$ around each datapoint $\mathbf{x}$.

\vspace{-0.08cm}
\begin{equation}
\label{def:fidelity}
\ell = \frac{1}{|{\mathcal X}|}\sum_{\mathbf{x}\in \mathcal X} \sum_{n \in {\mathcal N}_{\mathbf{x}}} F(m(n),\hat{c}_{\mathbf{x}}(n))
\end{equation}
\vspace{-0.08cm}

$\hat{c}_{\mathbf{x}}(n)$ is typically defined as something analogous to a first-order Taylor expansion about $\mathbf{x}$, taking the form $\hat{c}_\mathbf{x}(n) = m(\mathbf{x})+I\cdot(n-\mathbf{x})$, where instead of $I$ being the gradient of function $m$, it is the per-datapoint and per-feature importance returned by explainer $\mathcal{E}$.

However, as per datapoint feature importance is typically computed by fitting a simple linear model over the synthetic points \cite{lime}, we can simply consider a new feature-importance explanation method given by the per-point minimizer, thereby matching the explanation evaluation metric exactly:

\vspace{-0.08cm}
\begin{equation}
\label{eq:minimizer}
I'(\mathbf{x}) = {\arg\min}_I \sum_{n \in {\mathcal N}_{\mathbf{x}}} F(m(n),m(\mathbf{x})+I\cdot(n-\mathbf{x}))
\end{equation}

By definition, this is an optimal minimizer of \eqref{def:fidelity}, and will perform best with respect to the metric. As trivial examples of this: When $F$ is the squared loss, if $\mathcal{N}$ is defined in terms of homogeneous Gaussian noise then \eqref{eq:minimizer} corresponds to the definition of LIME \cite{disagreement_new}; as the variance of the Gaussian tends to 0, it corresponds to the gradient of the function; and it corresponds to SHAP, if $\mathcal{N}$ is chosen as weighted sampling over the vertices of a cube formed by swapping the values of a particular datapoint $p$ with the distribution mean.

Existing sensitivity based metrics such as PGI define \(F\) using \( L_1 \) loss, with neighborhood \(\mathcal{N}\) defined using a Gaussian distribution around datapoint \(\mathbf{x}\). While the loss is \(L_1\) and not \(L_2\), this formulation is otherwise interchangeable with LIME, and also converges to the gradient as the variance of the Gaussian tends to 0. This naturally promotes explainers that satisfy this definition of fidelity and neighborhood, \emph{violating the on-manifold evaluation principle}.

\vspace{-0.09cm}
\section{Methodology}
\vspace{-0.06cm}
\label{sec:methodology}

Inspired by previous work (section \ref{subsec:prior}) and desiderata from user studies \cite{predictions_for_explanations, user_predictiveness_1, user_predictiveness_2, user_predictiveness_3, user_predictiveness_4, user_predictiveness_5}, we consider a simple alternative to sensitivity-driven methods of XAI evaluation: the important features in any explanation should be more predictive of the model output than the unimportant features. \AXE adopts classifier accuracy \cite{axe_in_images} to measure the predictiveness of the top-n important features. This ``top-n'' style formulation, just like prior metrics from table \ref{tab:prior_metrics}, is considered intuitive for practitioners \cite{agarwal2022openxai}. 

For datapoint $\mathbf{x}$ and explanation $\mathbf{e}$, the top-n most important features are the importances with the largest absolute values. To measure explanation quality $q$, \AXE uses predictiveness -- the accuracy of a k-Nearest Neighbors (\textit{k}-NN) model $M^k$ in recovering the model prediction $m(\mathbf{x})$ using only the subset of the the top-n most important features. The \textit{k}-NN $M$ mimics the prediction $m(\mathbf{x})$ of model $m$ by averaging over the predictions from the \textit{k} neighbors nearest to $\mathbf{x}$ \cite{knn1, knn2}. We motivate our choice of \textit{k}-NN based on feature separability in section \ref{subsec:demotivation}.

\vspace{-0.04cm}
\subsection{The \AXE Framework}
\label{subsec:framework}

\vspace{-0.19cm}
\begin{algorithm}[ht]
\caption{Evaluating Explanation Quality with $\bm{\text{\axe}_n^k}$}
\label{alg:axe}
\begin{algorithmic}[1]
\REQUIRE Number of Features $n$, Number of Neighbors $k$ \\ Dataset $\mathcal{X} = \{\mathbf{x}_i\}_{i=1}^{\nu}$, Predictions $Y_\text{preds} = \{y_i\}_{i=1}^{\nu}$, and Explanations $E = \{\mathbf{e}_i\}_{i=1}^{\nu}$
\STATE Initialize an empty list: $\hat{Y} \gets []$
\FOR{each datapoint $\mathbf{x}_i$ and explanation $\mathbf{e}_i$ in $(\mathcal{X}, E)$}
    \STATE Find $n$ most important features: $f_\text{imp} \gets \text{ImpFeatures}(\mathbf{e}_i, n)$
    \STATE Create $\mathcal{X}_f$ with subset of features $f_{imp}$ from $\mathcal{X}$
    \STATE Train K-NN model $M^k_i$ with inputs $\mathcal{X}_f$ and target $Y_\text{preds}$
    \STATE Obtain prediction $\hat{y}_i$ from $M^k_i$ for datapoint $\mathbf{x}_i$
    \STATE Append $\hat{y}_i$ to $\hat{Y}$
\ENDFOR
\STATE Return performance measure: $ \text{Accuracy}(\hat{Y}, Y_\text{preds})$
\end{algorithmic}
\end{algorithm}
\vspace{-0.2cm}

Algorithm \ref{alg:axe} summarizes our framework for evaluating explanations. We denote the target variable \(Y\), predicted from the input dataset \(\mathcal{X}\) consisting of $\nu$ datapoints and $N$ features. The model \(m\) makes predictions \(Y_\text{preds} = m(\mathcal{X})\), for which a set of feature-importance explanations \(E\) can be computed such that $\exists \mathbf{e}_i \in E  \forall  \mathbf{x}_i \in \mathcal{X}$. \AXE fits multiple \textit{k}-NN models $M^k_i$ to predict model outputs $Y_\text{preds}$, not data labels $Y$. \AXE has two hyperparameters: \(n\) for the ``top-n'' number of important features to use and \(k\), for the number of neighbors to use in the \textit{k}-NN model $M^k$; denoted \(\text{\axe}_n^k \).

It is essential \emph{not} to use the same \textit{k}-NN model for all predictions -- \AXE does not build a global \textit{k}-NN surrogate to measure explanation quality. For each datapoint $\mathbf{x}_i$, we use a unique \textit{k}-NN model that considers the top-n most important features for that particular explanation $\mathbf{e}_i$. This insight is critical to ensure that \AXE does not just report the accuracy of an arbitrary \textit{k}-NN model over the entire dataset, and properly reflects the desiderata from the \emph{local contextualization} principle. By using \AXE to predict $Y_\text{preds}$ instead of $Y$, we satisfy the \emph{model relativism}, and by using the same training data as $m$, we satisfy \emph{on-manifold evaluation}.

For efficiency, the \textit{k}-NN models \(M_i^k\) trained in line 5 can be cached and reused. For a dataset with \( \nu \) features, the number of unique \textit{k}-NN models trained to compute \(\text{\axe}_k^n \) is at most \( \binom{N}{n} \). The cache size is bound by $\min (\nu, \binom{N}{n})$. Lastly, \AXE is flexible to allow the use of different performance measures in line 9. In our experiments we use accuracy:
\(
\text{Accuracy}(\hat{Y}, Y_{\text{preds}}) = \frac{1}{\nu} \sum_{i=1}^{\nu} \mathbf{1}_{(\hat{y}_{i} = y_i)}
\)

Instead of selecting a specific value of the hyperparameter \(n\), algorithm \ref{alg:axe} can be repeated for all possible values $n \in (0, N]$, computing multiple \(\text{\axe}_k^n\) scores. Finally, the area under a \(n\) -- \(\text{\axe}_n^k\) curve (AUC) can be used to obtain a single number as an overall evaluation of the model explanation quality, independent of $n$. This AUC trick is adopted from the literature and is a common way to obtain scalar scores from top-n based evaluation metrics \cite{agarwal2022openxai, abacus_xai_bench, Meng2022, 9150813}. This way, AXE can be made sensitive to the entire order of feature importances instead of just the top-n.

\AXE satisfies the \emph{local contextualization} principle by using a different set of neighbors for each datapoint $\mathbf{x}$. Each datapoint has its own nearest-neighbors model \(M_i^k\), unique to each prediction \(\hat{y}_i\) for $\mathbf{x}_i$. It satisfies the \emph{model relativism} principle by training model \(M_i^k\) to predict the classifier response \(Y_\text{preds}\), rather than the target feature \(Y\), making the quality metric $q$ dependent on the model $m$. Finally, \AXE satisfies the \emph{on-manifold evaluation} principle because the \textit{k}-NN models are explicitly limited to the existing data manifold and do not rely on new datapoints $\mathbf{x} \notin \mathcal{X}$, avoiding feature sensitivity measures. We can use \AXE to determine the quality scores for the explanations in figure \ref{fig:disagreement_example}. Using $\text{\axe}_{n=4}^{k=5}$, we get:- (a) Gradients: 0.4; (b) SHAP: 1.0; (c) LIME: 0.6; and (d) Integrated Gradients: 0.8, indicating SHAP is the most useful for predicting model behavior.

\vspace{-0.35cm}
\subsection{Illustrative Example}
\label{subsec:demotivation}

\begin{figure}[ht]
    \centering
    \vspace{-0.15cm}
    \includegraphics[width=0.85\linewidth]{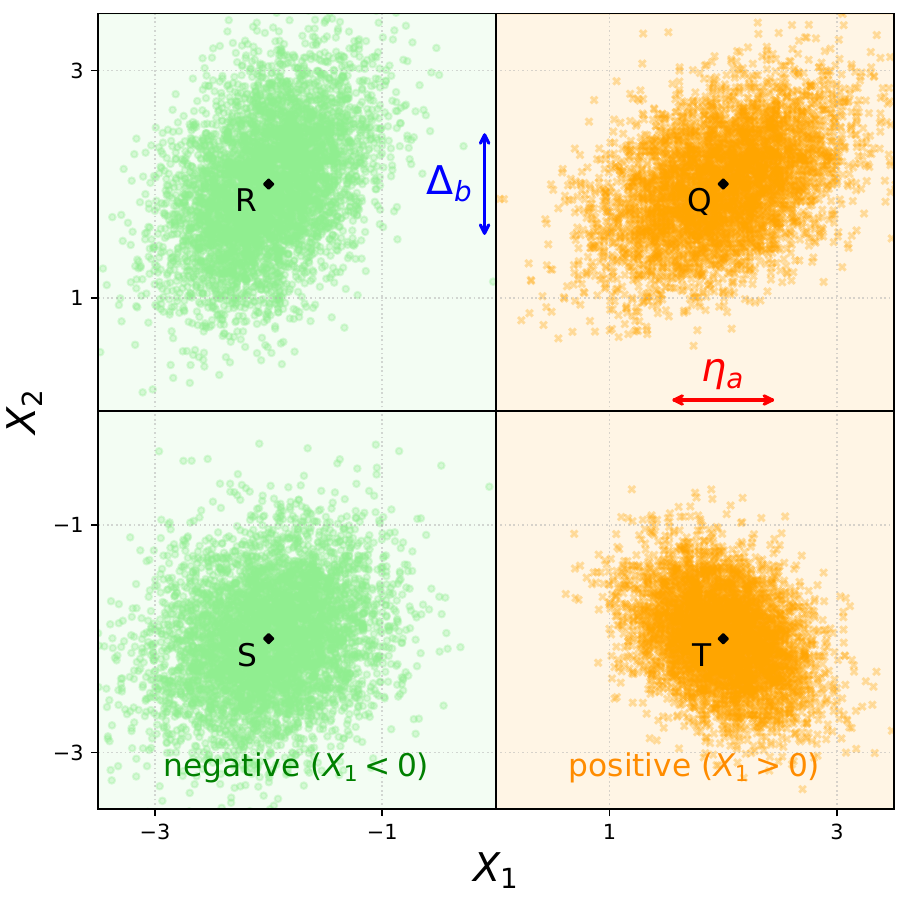}
    \caption{\textbf{Synthetic Data and Model for AXE and PGI evaluations:} 4 Normal distributions representing the data distribution, and neighborhoods $\Delta$ and $\eta$ for PGI and \AXE respectively. The model is defined as $m(\mathbf{x}) = \mathbf{1}_{X_1 > 0}$, and we compare the quality of competing explanations $\mathbf{e}_a$ ($X_1$ is more important) and $\mathbf{e}_b$ ($X_2$ is more important) for datapoint $Q$.}
    \label{fig:motivation}
    \vspace{-0.15cm}
    \Description[]{}
\end{figure}

Intuitively, \AXE uses \textit{k}-NN models because we want important features to be those that separate model predictions in feature space. Conversely, unimportant inputs should be unable to separate the model predictions from each other. To illustrate the intuition behind the choice of \textit{k}-NN models in \AXE, we present a motivating example. Consider a dataset consisting of input features $X_1$ and $X_2$, sampled from 4 Normal distributions illustrated in figure \ref{fig:motivation}.

The data (5000 points each) is sampled from 4 Normal distributions $\mathcal{N}_Q$, $\mathcal{N}_R$, $\mathcal{N}_S$, and $\mathcal{N}_T$, with varying covariances, respectively centered at $Q=(2,2)$, $R=(-2,2)$, $S=(-2,-2)$, and $T=(2,-2)$. Model $m$ makes predictions using only input feature $X_1$, independent of feature $X_2$. By construction, for all non-outliers we can assume: $m(\mathbf{x}_q)=1 \forall \mathbf{x}_q \sim \mathcal{N}_Q$, $m(\mathbf{x}_r)=0 \forall \mathbf{x}_r \sim \mathcal{N}_R$, $m(\mathbf{x}_s)=0 \forall \mathbf{x}_s \sim \mathcal{N}_S$, and $m(\mathbf{x}_t)=1 \forall \mathbf{x}_t \sim \mathcal{N}_T$. Explanation $\mathbf{e}_a = [i_{a1}, i_{a2}]$ has $i_{a1} > i_{a2}$ ($X_1$ is more important), and explanation $\mathbf{e}_b = [i_{b1}, i_{b2}]$ has $i_{b1} < i_{b2}$ ($X_2$ is more important). Ideally, metric $q$ should correctly assign a higher score to an explanation $\mathbf{e}_a$: $q(\mathbf{e}_a) > q(\mathbf{e}_b)$. For $\mathbf{e}_a$ (and $\mathbf{e}_b$), the PGI perturbation neighborhood is $\Delta_a$ (and $\Delta_b$) and the AXE \textit{k}-NN neighborhood is $\eta_a$ (and $\eta_b$) along the respective axes $X_1$ (and $X_2$) respectively. In practice, the $\Delta$ and $\eta$ neighborhoods are very similar.

\begin{figure*}[htb]
    \centering
    \subfloat[\textbf{\axe}: \AXE relaibly shows that explanation $\mathbf{e}_a$ is better than explanation $\mathbf{e}_b$. $\text{\AXE}_{n=1}(\mathbf{e}_a) > \text{\AXE}_{n=1}(\mathbf{e}_b) \forall k \in (1,10000)$.]{
        \includegraphics[width=0.48\linewidth]{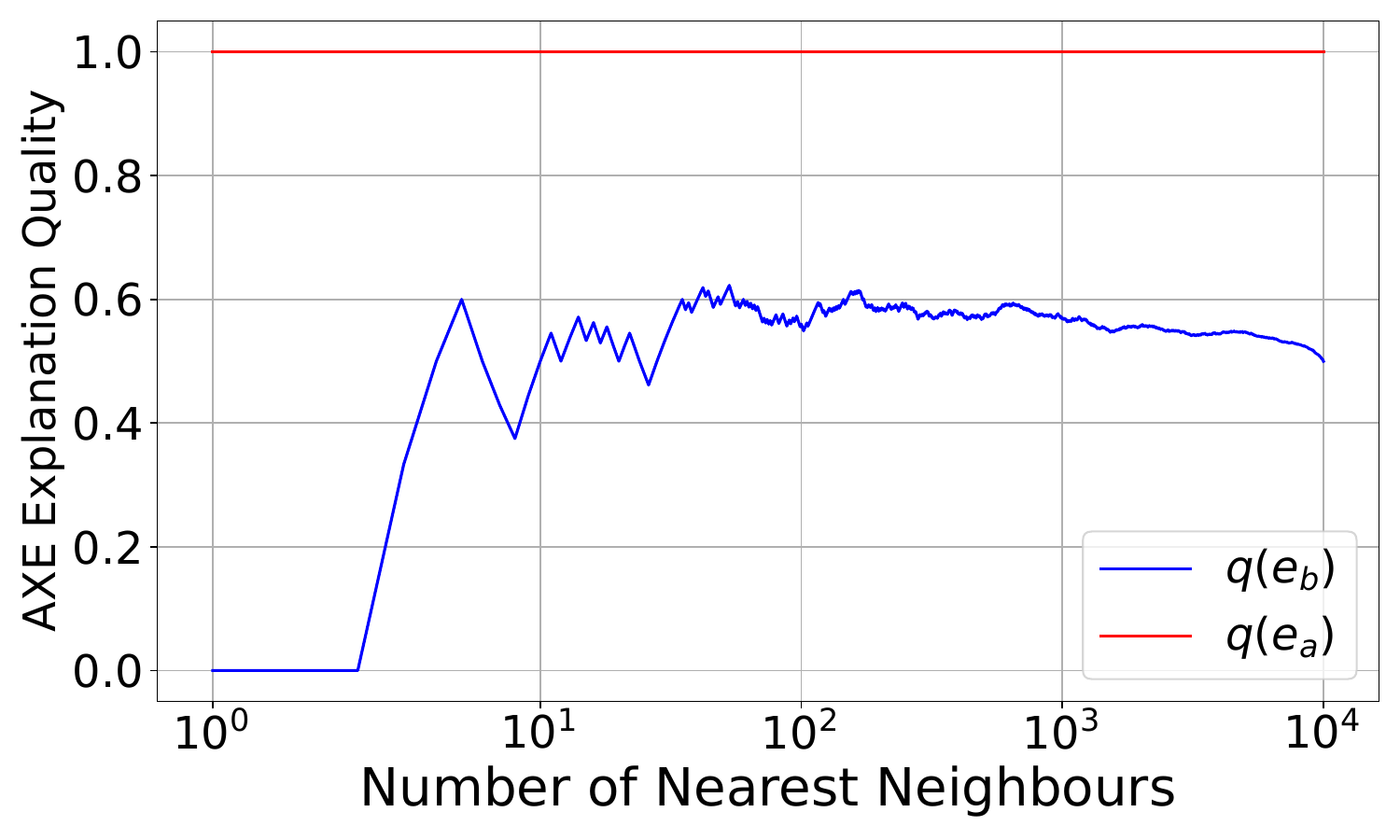}
    }
    \hfill
    \subfloat[\textbf{PGI}: $\text{PGI}_{n=1}(\mathbf{e}_a) > \text{PGI}_{n=1}(\mathbf{e}_b)$ only once the perturbations are large, where on-manifold probability $P(\Delta \subseteq \mathcal{M})$ is low.]{
        \includegraphics[width=0.48\linewidth]{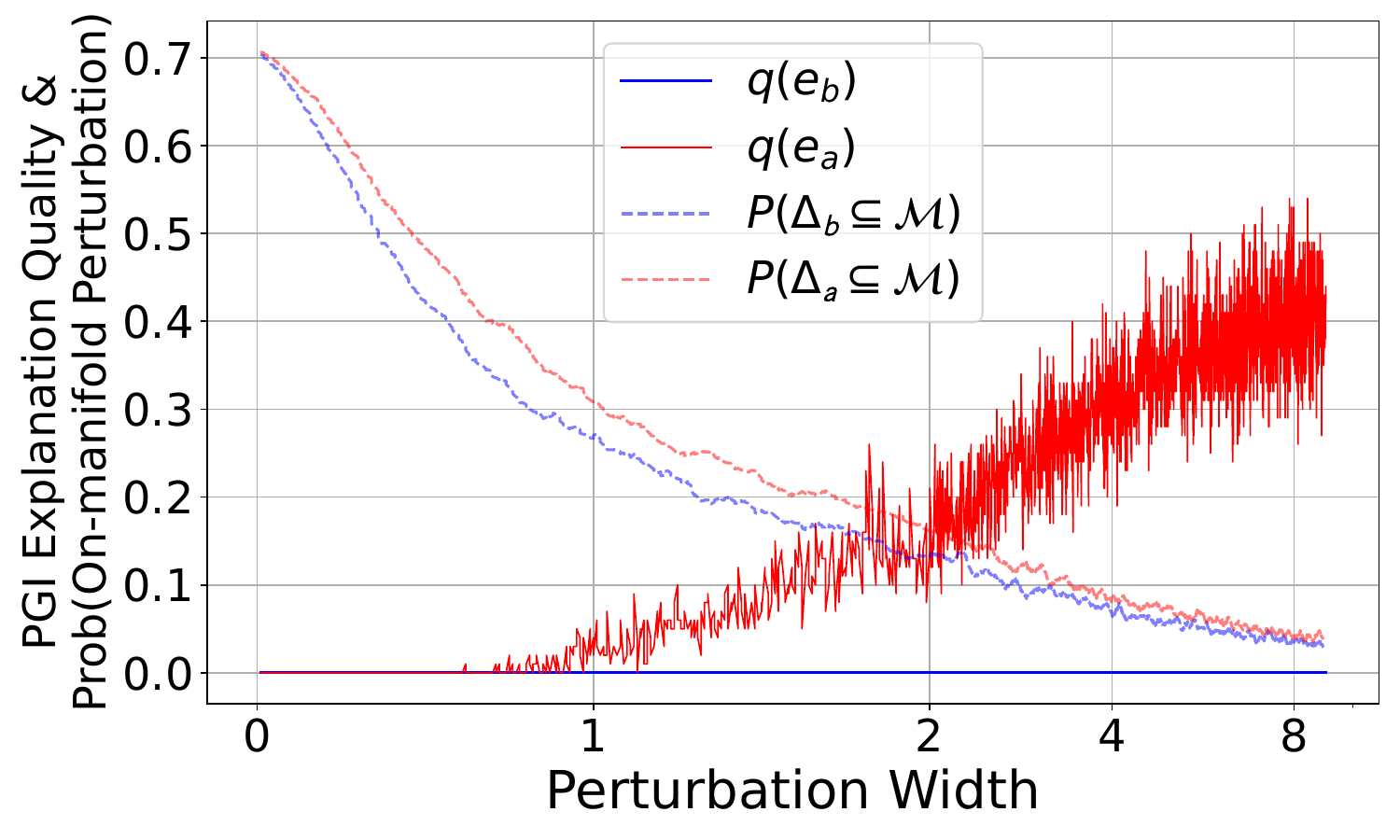}
    }
    \vspace{-0.05cm}
    \caption{\textbf{Comparing explanations using \AXE and PGI}: By definition $q(\mathbf{e}_a) > q(\mathbf{e}_b)$, but PGI does not clearly show this. \AXE correctly determines that explanation $\mathbf{e}_a$ is better than $\mathbf{e}_b$, across hyperparameter values. (Both X axes on symlog scale).}
    \label{fig:axe_v_pgi}
    \Description[]{}
    \vspace{-0.1cm}
\end{figure*}

We restrict our illustrative analysis to explanations for a single datapoint $Q=(2,2)$. From the definition of model $m$, it is clear that $X_1$ should be more important that $X_2$, and consequently $\mathbf{e}_a$ better than $\mathbf{e}_b$. Upon examination, this can easily be verified to be the case for $\text{\axe}_{n=1}$. Consider $\mathbf{e}_b$ where $X_2$ is the more important feature. A nearest neighbor model finding neighbors for $Q$ considering only $X_2$ and ignoring $X_1$ (per the \AXE definition) will find datapoints in the \textit{k}-NN neighborhood $\eta_b$ for point $Q$ and feature $X_2$. This would include points $\mathbf{x}_q$, and $\mathbf{x}_r$, but not $\mathbf{x}_s$ or $\mathbf{x}_t$. Predictions from these points can be either 1 or 0 respectively, hence the nearest neighbor model will have both labels in its neighbors, predicting an average near 0.5. This implies poor accuracy in recovering the positive prediction $m(Q)$, leading to a low \AXE score ($\sim 0.5$, see figure \ref{fig:axe_v_pgi} a). On the other hand, for explanation $\mathbf{e}_a$, since $X_1$ is the important feature, the neighborhood $\eta_a$ will include points from $\mathbf{x}_q$ and $\mathbf{x}_t$. These are all predicted to fall in the positive class, thus recovering $m(Q)$ with perfect accuracy of 1.0 and leading to a high \AXE score ($\sim 1.0$, see figure \ref{fig:axe_v_pgi} a). Figure \ref{fig:axe_v_pgi} (a) plots $\text{AXE}_{n=1}(\mathbf{e}_a)$ and $\text{AXE}_{n=1}(\mathbf{e}_b)$ for different $k$ values for the \textit{k}-NN models in \axe, clearly showing $\text{\AXE}_{n}(\mathbf{e}_a) > \text{\AXE}_{n}(\mathbf{e}_b)$ for all hyperparameter values of $k$.

We now analyze the behavior of PGI. For $\mathbf{e}_b$, PGI would generate datapoints $(2.0,2.0+\delta), \delta \sim \mathcal{N}(0,\text{width})$. Varying $X_2$ has no impact on the model prediction (by definition), yielding a prediction gap of 0. Conversely, for $\mathbf{e}_a$, PGI would sample datapoints $(2.0+\delta,2.0), \delta \sim \mathcal{N}(0,\text{width})$. The predictions for these are highly sensitive to the neighborhood $\Delta_a$, the the PGI sampling width. This hyperparameter determines whether the neighborhood stays on the same side of the decision boundary $X_1 = 0$. If it does, then PGI is 0 -- a result that provides no information to compare $\mathbf{e}_a$ and $\mathbf{e}_b$. Figure \ref{fig:axe_v_pgi} (b) shows that PGI is zero until the neighborhood becomes large enough. However large neighborhoods present a different challenge -- the points PGI samples are more likely to lie off manifold. Figure \ref{fig:axe_v_pgi} (b) also shows the corresponding probability of the PGI perturbations lying on manifold, and it can be seen that in the "useful" non-zero range of the plot, on-manifold probabilities are lower. In general, it is difficult to tune the neighborhood width hyperparameter in sensitivity analysis \cite{molnar_book, Riley2023-rs}. Tuning this hyperparameter requires knowing apriori what explanations to expect -- an implausible expectation akin to knowing ``ground-truth'' explanations. Lastly, figure \ref{fig:axe_v_pgi} (b) shows that PGI is unstable even in regions of high perturbation width, further complicating its use in practice.

\AXE does not require model predictions on off-manifold data. \AXE also does not need access to ground-truth explanations. This is enabled by using \textit{k}-NN model accuracy to measure explanation quality -- fitting a different \textit{k}-NN model for each unique explanation and datapoint. Critically, using \textit{k}-NN models provides a number of advantages. \textit{k}-NN models operate on the classifier data itself, omitting the need for off-manifold predictions, satisfying the \emph{on-manifold evaluation} principle. Further, the choice of nearest neighbor models directly captures the notion of separability in feature space -- capturing the idea that an important feature is one that separates classes in feature space.

\vspace{-0.1cm}
\section{Experiments}
\label{sec:experiments}

In sections \ref{subsec:gt_failures}, \ref{subsec:sens_failures}, and \ref{subsec:demotivation} we used synthetic examples demonstrating the ineffectiveness of prior explanation evaluation metrics. We now compare \AXE with existing baselines on real-world datasets.

\subsection{Detecting Explanation Fairwashing}
\label{subsec:adv_attack}

\begin{table*}[ht!]
\caption{
\textbf{Detecting explanation fairwashing}: We replicate an adversarial fairwashing attack \cite{adv_attack} and generate spurious explanations, which we try to then detect using quality metrics $q$ that do not need ground-truths: \AXE, PGI, and PGU. Complete details are in section \ref{subsec:adv_attack}: $E_\rho$ is a set of explanations that correctly denote that the most important model input feature is $X_\rho$. $E_\phi$ and $E_\psi$ are sets of manipulated explanations created by an adversary where the most important feature is $X_\phi$ or $X_\psi$. \\
A good evaluation metric $q$ should distinguish manipulated explanations from correct explanations -- we expect that: (i) \(\bar{q}(E_\rho) > \bar{q}(E_\phi) \) and (ii) \(\bar{q}(E_\rho) > \bar{q}(E_\psi) \); where $\bar{q}(E) = \sum_{\mathbf{e} \in E} q(\mathbf{e}) / |E|$. Cases where only one of the two conditions is a strict inequality are marked with an asterisk*. \textbf{\AXE has a success rate of 100\%}, whereas the \textbf{overall success rate for PGI and PGU is only 50\%}.
}
\vspace{-0.6cm}
\label{tab:pgiufails}
\centering
\begin{tabular}{ll| @{\hspace{1.5em}} r @{\hspace{1.5em}} r @{\hspace{2em}} r @{\hspace{2em}} r r @{\hspace{1.5em}} |c}
& & & & & & & \\
\toprule
\multirow{2}{*}{\thead{ \\ \textbf{Dataset} }} & 
\multirow{2}{*}{\thead{ \textbf{Adversarial} \\ \textbf{Model} \\  \(\bm{m_L}\) \textit{or} \(\bm{m_S}\) }} & 
\multirow{2}{*}{ \thead{ \textbf{Eval.} \\ \textbf{Metric $q$} \\ (\(n=1\)) }} & 
\multicolumn{4}{l|}{\thead{\textit{Evaluating explanations with a single important attribute:}}} & 
\multirow{2}{*}{\thead{  \bm{$q(E_\rho) > q(E_\phi)$} \\ \textit{and} \\ \bm{$q(E_\rho) > q(E_\psi)$} }} \\

 & & & \thead{\textbf{Protected} \\ \bm{$q(E_\rho)$} } & 
 \thead{\textbf{Foil 1} \\ \bm{$q(E_\phi)$} } & 
 \thead{\textbf{Foil 2} \\ \bm{$q(E_\psi)$} } & 
 \thead{\textbf{Other} \\ \bm{$q(E_\omega)$} } \\

\specialrule{\lightrulewidth}{0.2em}{0pt}

\multirow{6}{*}{\makecell[l]{ \textbf{German} \\ \textbf{Credit} } }
 & \multirow{3}{*}{ \makecell[l]{ \(\bm{m_L}\) \\ \textbf{\textit{(1 foil)}} } } 

 & PGI & 0.032 & 0.148 & na & 0.018 & \incorrect \\
 & & (-)PGU & -0.486 & -0.536 & na & -0.483 & \correct \\ 
 & & \axe & 1.000 & 0.680 & na & 0.617 & \correct \\ 

\cdashline{2-8}[1pt/2pt]
 & \multirow{3}{*}{ \makecell[l]{ \(\bm{m_S}\) \\ \textbf{\textit{(1 foil)}} } } 

 & PGI & 0.037 & 0 & na & 0.037 & \correct \\
 & & (-)PGU & -0.475 & -0.529 & na & -0.478 & \correct \\
 & & \axe & 0.990 & 0.690 & na & 0.622 & \correct \\

\hdashline[5pt/2pt]

\multirow{12}{*}{ \makecell[l]{ \textbf{COMPAS} } } 
 & \multirow{3}{*}{ \makecell[l]{ \(\bm{m_L}\) \\ \textbf{\textit{(1 foil)}} } } 

 & PGI & 0.006 & 0 & na & 0.067 & \correct \\
 & & (-)PGU & -0.481 & -0.479 & na & -0.431 & \incorrect \\ 
 & & \axe & 0.992 & 0.739 & na & 0.534 & \correct \\ 

\cdashline{2-8}[1pt/2pt]
 & \multirow{3}{*}{ \makecell[l]{ \(\bm{m_S}\) \\ \textbf{\textit{(1 foil)}} } } 

 & PGI & 0.006 & 0.035 & na & 0.009 & \incorrect \\
 & & (-)PGU & -0.091 & -0.077 & na & -0.090 & \incorrect \\ 
 & & \axe & 0.968 & 0.761 & na & 0.527 & \correct \\ 

\cdashline{2-8}[3pt/2pt]
 & \multirow{3}{*}{ \makecell[l]{ \(\bm{m_L}\) \\ \textbf{\textit{(2 foils)}} } } 

 & PGI & 0.006 & 0 & 0.001 & 0.075 & \correct \\
 & & (-)PGU & -0.520 & -0.520 & 0-0.524 & -0.464 & \incorrect* \\
 & & \axe & 0.990 & 0.739 & 0.735 & 0.533 & \correct \\ 

\cdashline{2-8}[1pt/2pt]
 & \multirow{3}{*}{ \makecell[l]{ \(\bm{m_S}\) \\ \textbf{\textit{(2 foils)}} } } 

 & PGI & 0.005 & 0.039 & 0.041 & 0.010 & \incorrect \\
 & & (-)PGU & -0.104 & -0.090 & -0.092 & -0.106 & \incorrect \\
 & & \axe & 0.956 & 0.746 & 0.731 & 0.531 & \correct \\ 

 \hdashline[5pt/2pt]

\multirow{12}{*}{ \makecell[l]{ \textbf{Communities} \\ \textbf{and Crime} } } 
 & \multirow{3}{*}{ \makecell[l]{ \(\bm{m_L}\) \\ \textbf{\textit{(1 foil)}} } } 

 & PGI & 0.103 & 0 & na & 0.029 & \correct \\
 & & (-)PGU & -0.479 & -0.460 & na & -0.481 & \incorrect \\ 
 & & \axe & 1.000 & 0.765 & na & 0.793 & \correct \\ 

\cdashline{2-8}[1pt/2pt]
 & \multirow{3}{*}{ \makecell[l]{ \(\bm{m_S}\) \\ \textbf{\textit{(1 foil)}} } } 

 & PGI & 0.089 & 0.006 & na & 0.005 & \correct \\
 & & (-)PGU & -0.446 & -0.429 & na & -0.448 & \incorrect \\ 
 & & \axe & 0.985 & 0.765 & na & 0.790 & \correct \\

\cdashline{2-8}[5pt/2pt]
 & \multirow{3}{*}{ \makecell[l]{ \(\bm{m_L}\) \\ \textbf{\textit{(2 foils)}} } } 

 & PGI & 0.101 & 0.001 & 0.001 & 0.034 & \correct \\
 & & (-)PGU & -0.534 & -0.536 & -0.536 & -0.535 & \correct \\
 & & \axe & 0.995 & 0.760 & 0.760 & 0.792 & \correct \\

 \cdashline{2-8}[1pt/2pt]
 & \multirow{3}{*}{ \makecell[l]{ \(\bm{m_S}\) \\ \textbf{\textit{(2 foils)}} } } 

 & PGI & 0.094 & 0.006 & 0.005 & 0.008 & \correct \\
 & & (-)PGU & -0.479 & -0.470 & -0.479 & -0.479 & \incorrect* \\
 & & \axe & 0.955 & 0.760 & 0.755 & 0.781 & \correct \\ 

\specialrule{\heavyrulewidth}{0pt}{0pt}

\end{tabular}
\end{table*}

We simulate a state-of-the-art adversarial attack \cite{adv_attack} on explanations in a real-world setting where ground-truths remain unknown. The attack modifies a model $m$, known to be discriminatory, creating new models $m_S$ or $m_L$ that respectively fool SHAP and LIME into generating explanations $\mathbf{e}$ that show the discriminatory feature as unimportant. A good explanation evaluation metric $q$ should identify manipulated explanations by scoring them poorly.

This attack fairwashes model explanations by hiding discriminatory model behavior. Imagine explanations for the diabetes prediction model figure \ref{fig:disagreement_example} showing that the model used benign inputs to make its prediction, when the model actually made predictions using protected attributes that are medically irrelevant.

Like $m$, $m_S$ and $m_L$ too make decisions using only the ``protected'' feature ($X_\rho$), but they fool explainers SHAP and LIME respectively into generating explanations showing spurious "foil" features ($X_\phi, X_\psi$) as the most important \cite{adv_attack}. We use the same datasets and models as the original adversarial attack \cite{adv_attack}: the German Credit dataset from lending \cite{german_credit} and the COMPAS \cite{compas} dataset and the Communities and Crime dataset \cite{communities_and_crime} from criminal justice.

\begin{figure*}[htb]
    \centering
    \subfloat{
        \includegraphics[width=0.47\linewidth]{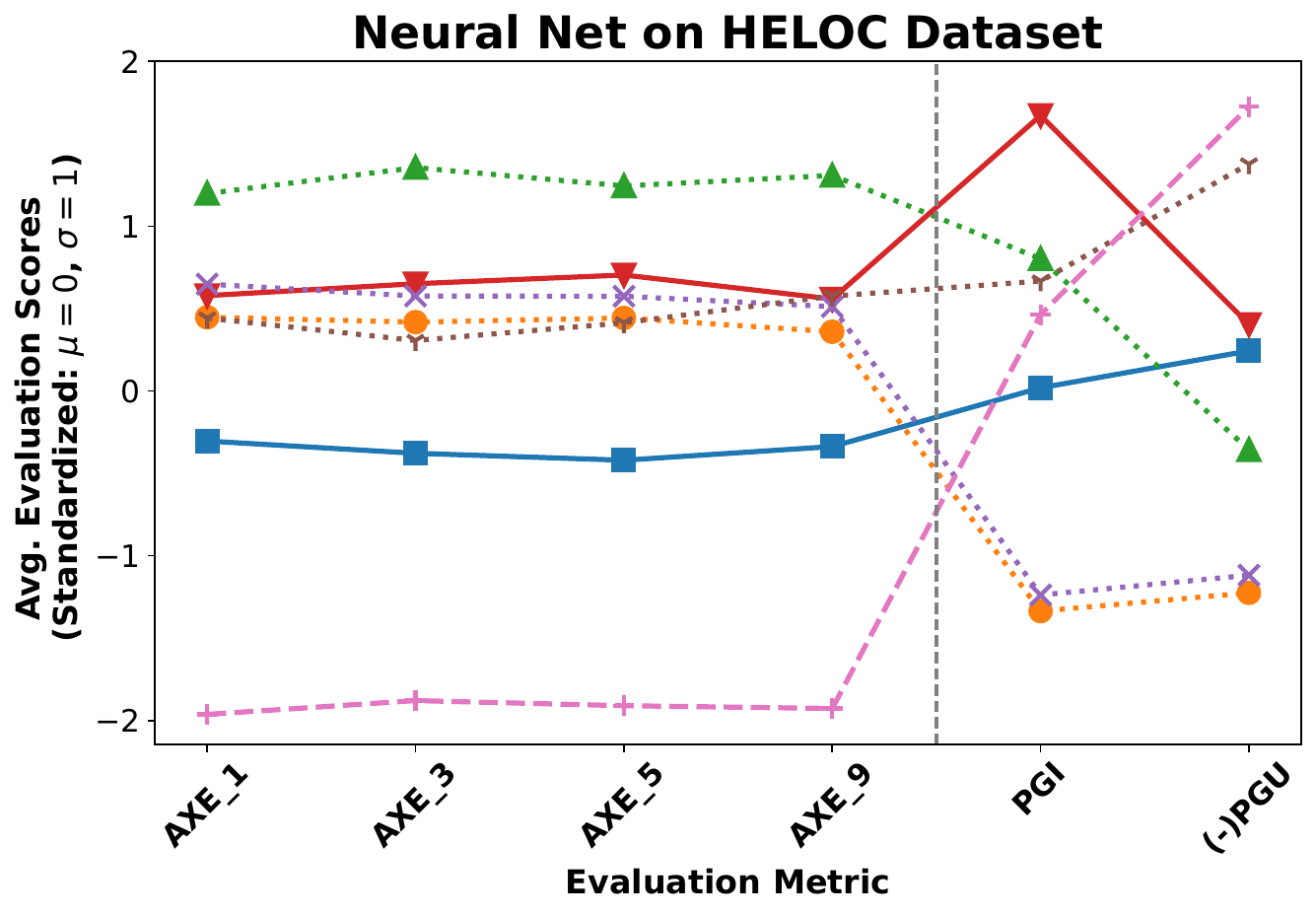}
    }
    \hfill
    \subfloat{
        \includegraphics[width=0.47\linewidth]{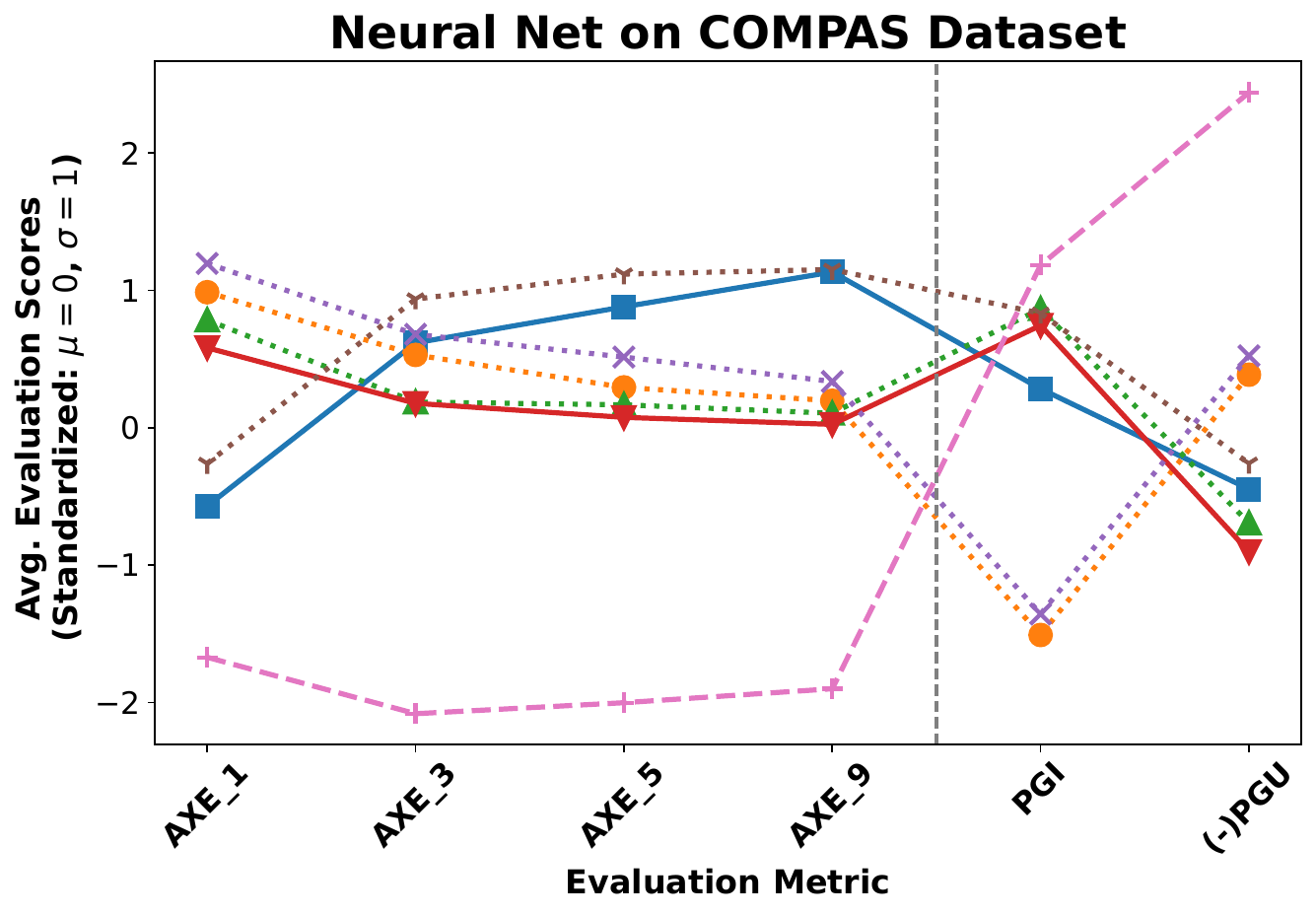}
    }
    \hfill
    \subfloat{
        \includegraphics[width=0.47\linewidth]{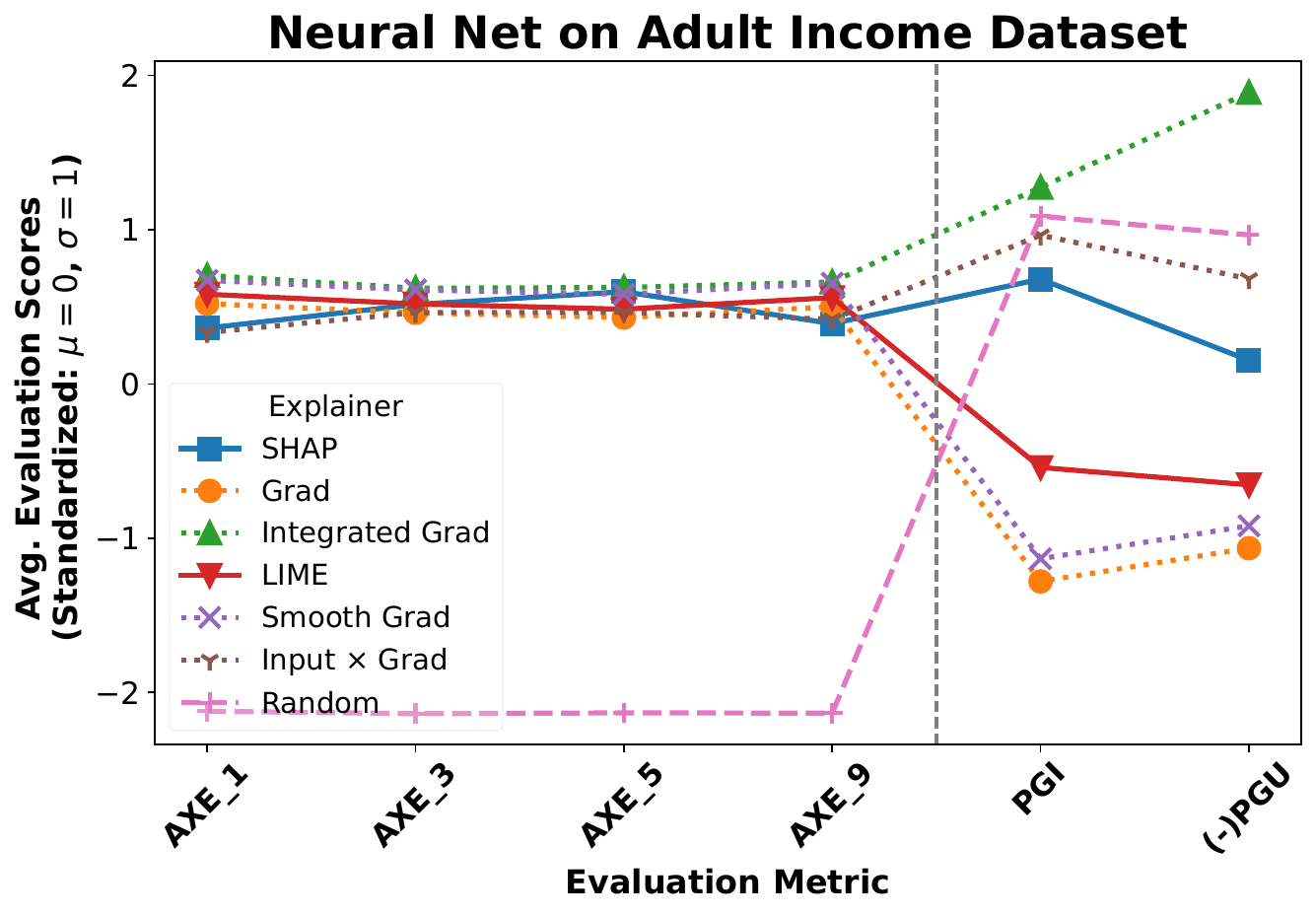}
    }
    \hfill
    \subfloat{
        \includegraphics[width=0.47\linewidth]{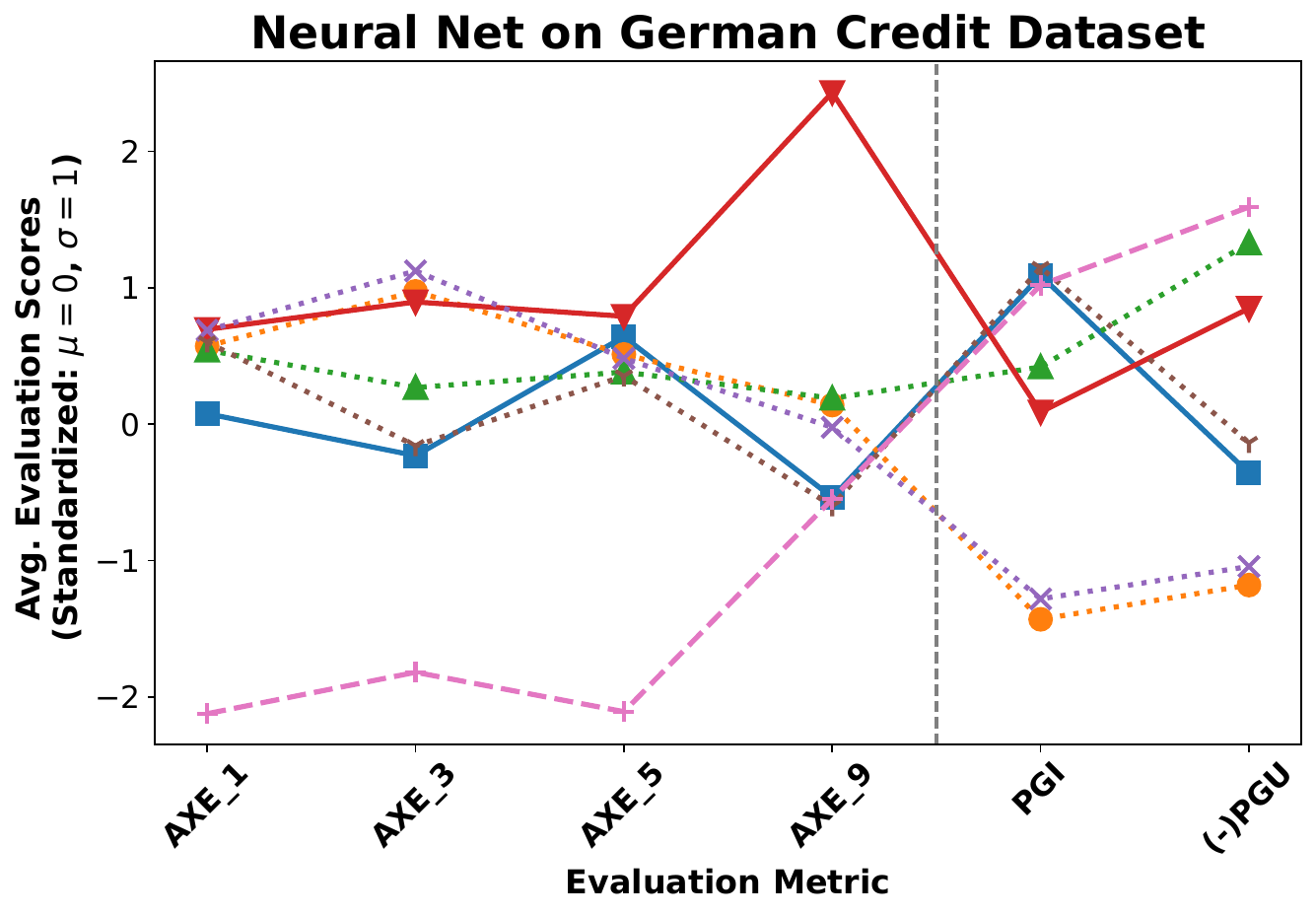}
    }
    \vspace{-0.1cm}
    \caption{\textbf{Evaluating the Quality of Explanations for Neural Networks:} For a fair comparison across evaluation metrics, we average over the entire dataset and plot their Z-score standardized values. Neural nets have no ground truth explanation $\mathbf{e}^{*}$, so the only metrics available are \axe, PGI, and -PGU (PGU inverted so higher values are better). Instead of a particular number of top-n features, we use the AUC trick from section \ref{subsec:framework}. For details see section \ref{subsec:openxai_experiments}.}
    \vspace{-0.1cm}
    \Description[]{}
    \label{fig:nn_openxai}
\end{figure*}

We then manually construct competing feature-importance explanations: $E_\rho$ is the set of all explanations where the protected feature $X_\rho$ is the most important feature. $E_{\phi}$ and $E_{\psi}$ are the explanation sets where the foil feature $X_\phi$ or $X_\psi$ is the most important, respectively. Finally, $E_\omega$ is the set of all other explanations, where the most important feature is neither protected nor a foil feature. The credit model is adversarially modified to deliberately favor men over others, and the criminal justice models are modified to favor white people over others. Explanations that reveal this ($E_\rho$) are correct. Explanations that mask this ($E_{\phi},E_{\psi}$) are spurious. Evaluation metrics $q$ should identify this by the condition $\bar{q}(E_{\rho}) > \bar{q}(E_{\phi})$ and $\bar{q}(E_{\rho}) > \bar{q}(E_{\psi})$; where $\bar{q}$ is the average quality $\bar{q}(E) = \sum_{\mathbf{e} \in E} q(\mathbf{e}) / |E|$.

In table \ref{tab:pgiufails} we summarize our results. We use \(\text{\AXE}_{n=1}\),  \(\text{PGI}_{n=1}\) and  \(\text{PGU}_{n=1}\) to measure the quality of the 3 sets of explanations. We fix top-n as 1 in each case because our explanations are constructed to only promote one feature as important at a time. As explained, our verification test for \AXE is that $\text{AXE}(E_\rho) > \text{AXE}(E_\phi)$ and $\text{AXE}(E_\rho) > \text{AXE}(E_\psi)$, which is found to always be true. Both PGI and PGU fail their corresponding checks.

The last column of table \ref{tab:pgiufails} shows PGU failing to discern genuine explanations $E_\rho$ from spurious ones $E_\psi$, $E_\phi$ 7 out of 10 times, and PGI failing to do so 3 out of 10 times. \AXE never fails, placing the overall error rate for sensitivity metrics PGI and PGU at 50\%, and for \AXE at 0\%. This indicates the evaluation metrics PGI and PGU are not impartial -- their optimization objectives are so aligned with LIME and SHAP that adversarial models designed to fool LIME and SHAP end up fooling PGI and PGU too. Table \ref{tab:pgiufails} proves experimentally that PGI and PGU violate the on-manifold evaluation principle, as we showed theoretically in section \ref{subsec:sens_failures}.

\subsection{Comparing \AXE with Baselines}
\label{subsec:openxai_experiments}

\begin{figure*}[htp]
    \centering
    \subfloat{
        \includegraphics[width=0.47\linewidth]{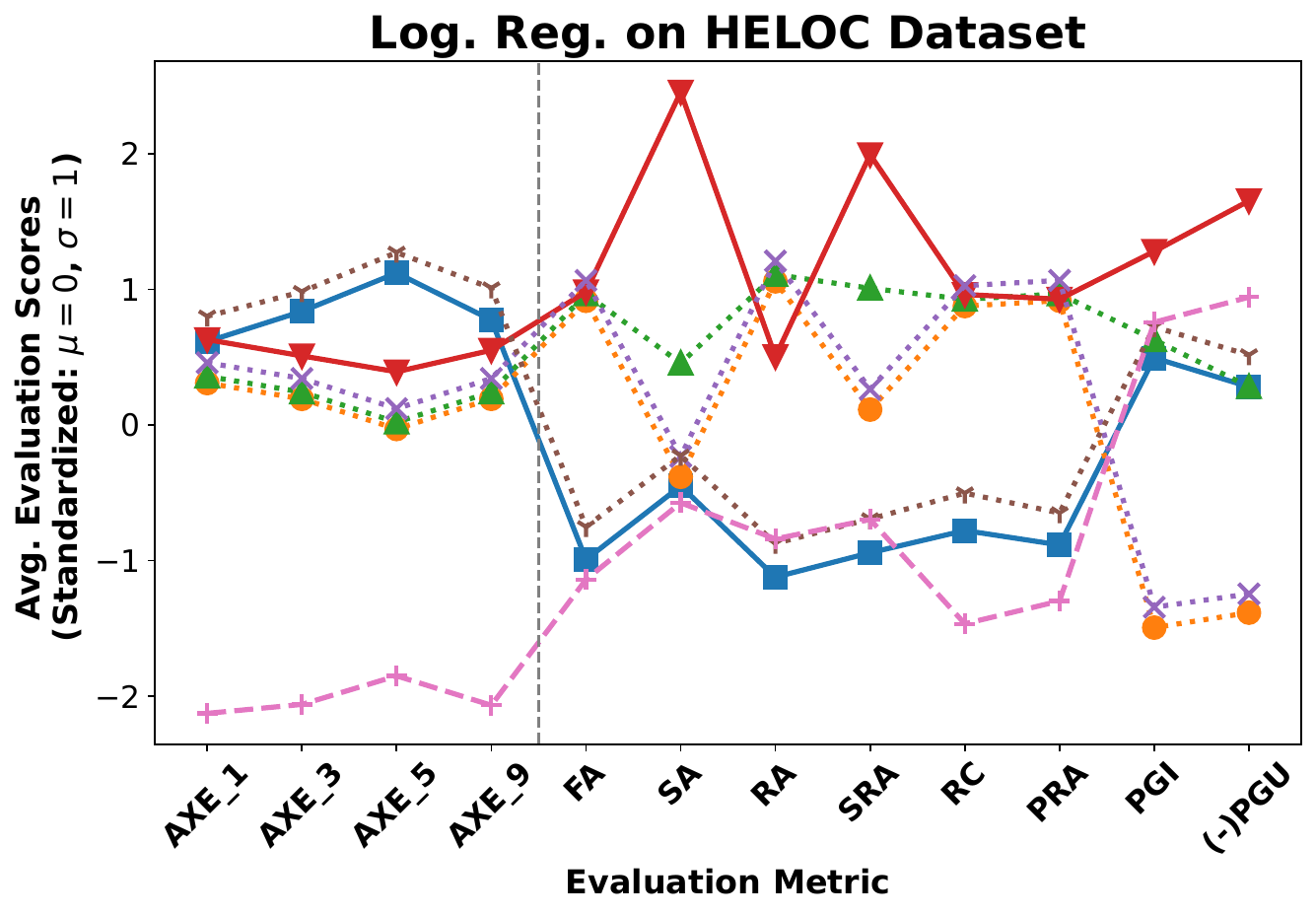}
    }
    \hfill
    \subfloat{
        \includegraphics[width=0.47\linewidth]{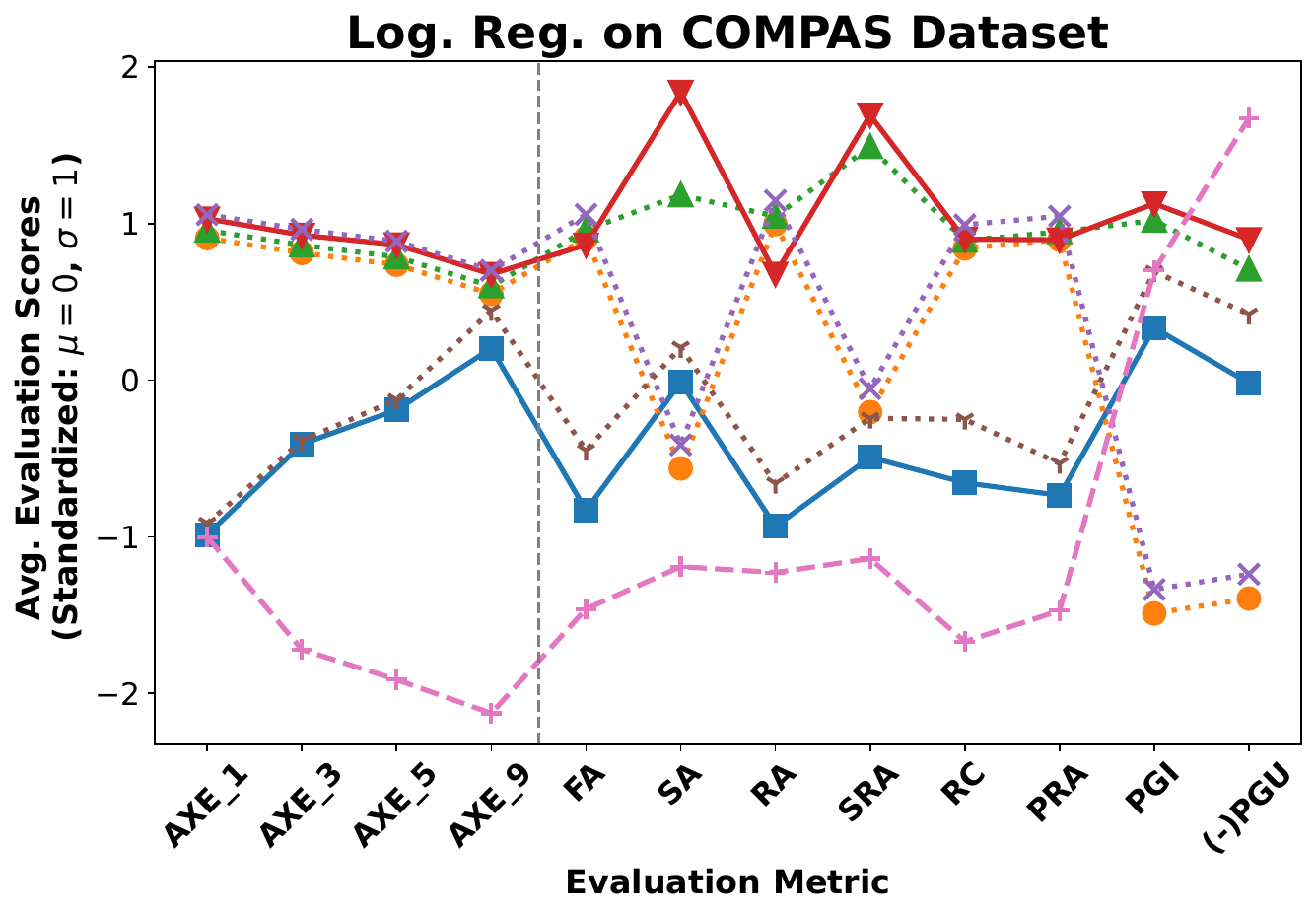}
    }
    \hfill
    \subfloat{
        \includegraphics[width=0.47\linewidth]{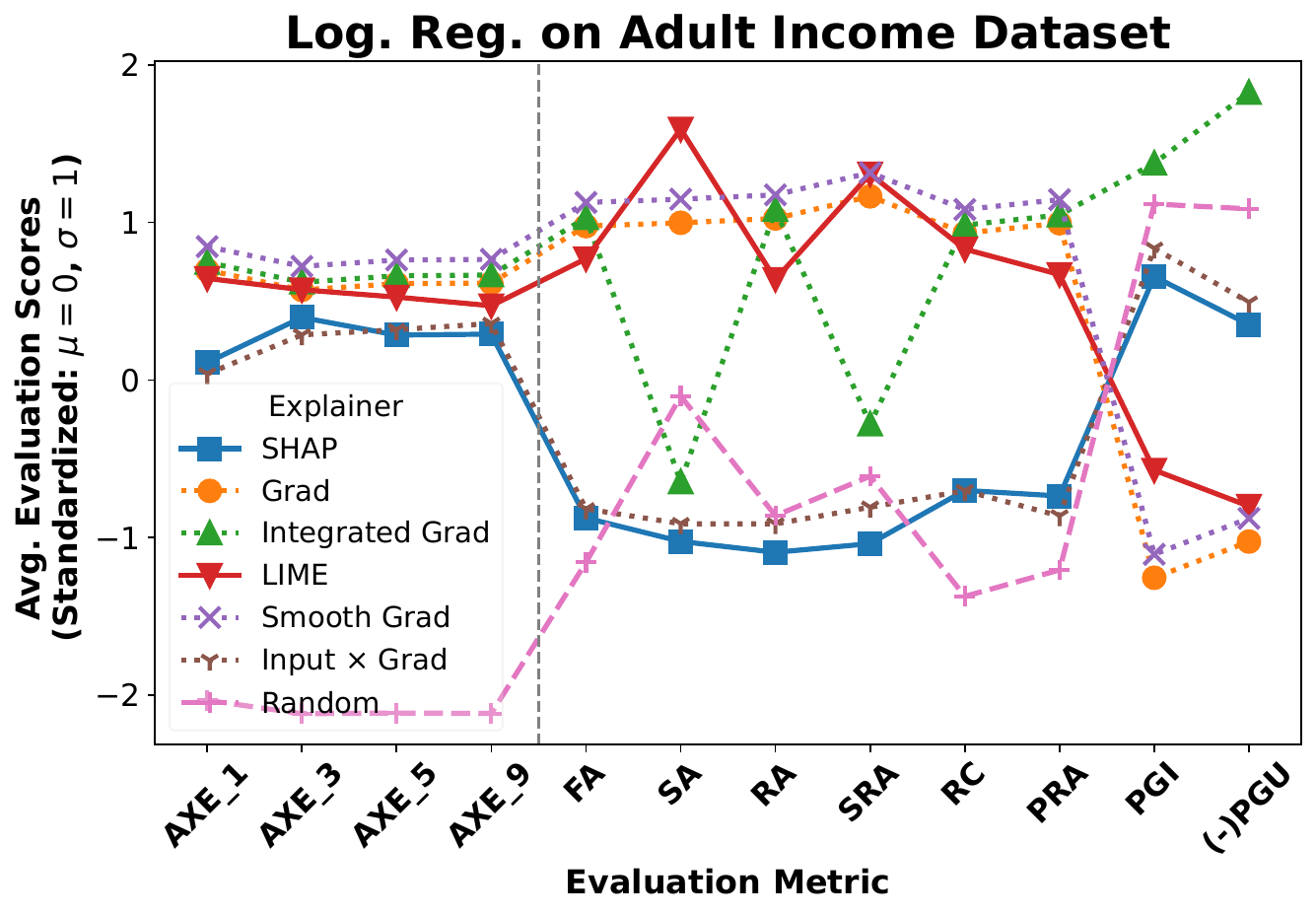}
    }
    \hfill
    \subfloat{
        \includegraphics[width=0.47\linewidth]{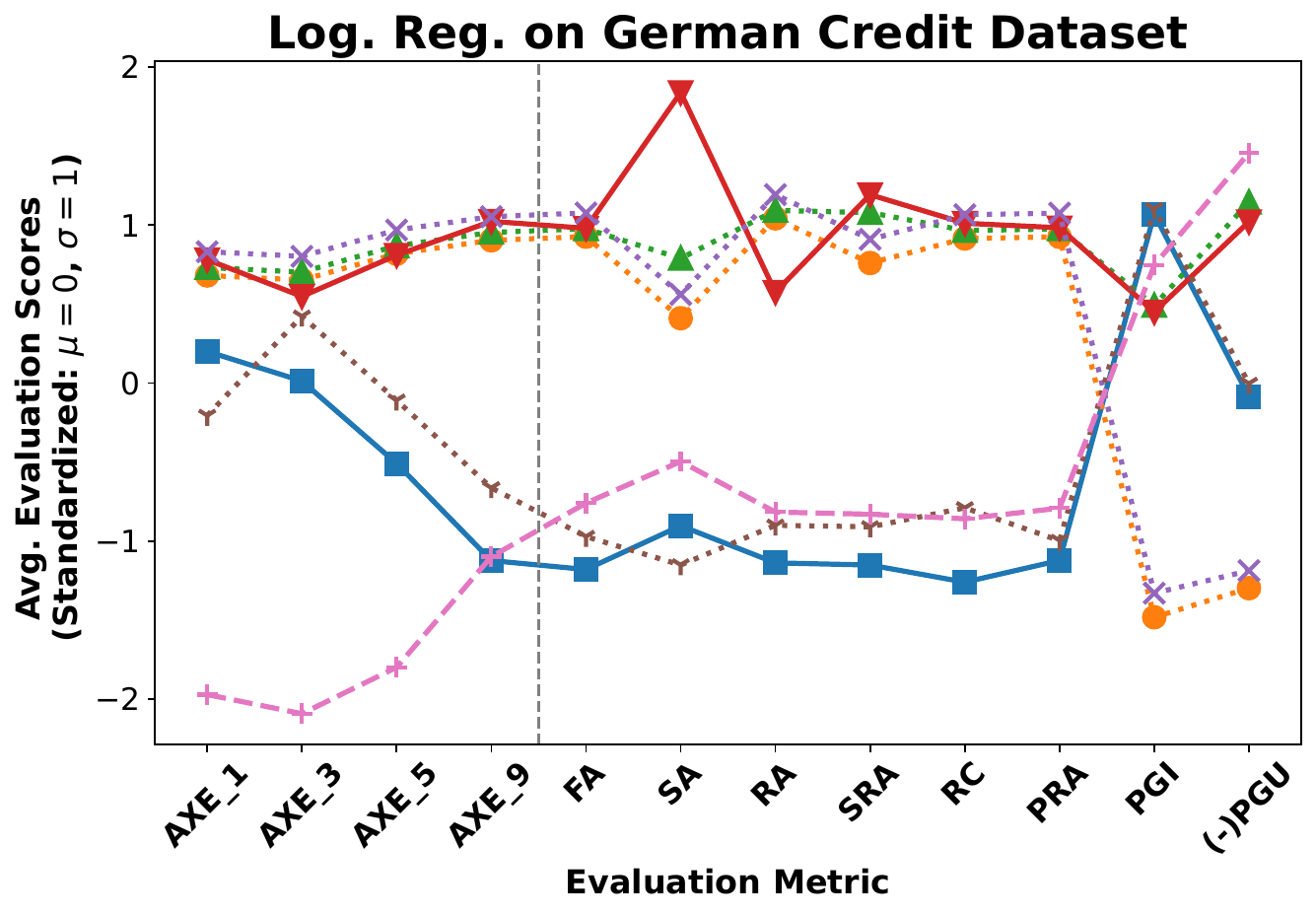}
    }
    \vspace{-0.1cm}
    \caption{\textbf{Evaluating the Quality of Explanations for Logistic Regression:} For a fair comparison across evaluation metrics, we average over the entire dataset and plot their Z-score standardized values. We compare \AXE with all prior metrics from table \ref{tab:prior_metrics}: FA, RA, SA, SRA, RC, PRA, PGI, and -PGU (PGU inverted so higher values are better). Instead of a particular number of top-n features, we use the AUC trick from section \ref{subsec:framework}. For details see section \ref{subsec:openxai_experiments}.}
    \vspace{-0.1cm}
    \Description[]{}
    \label{fig:lr_openxai}
\end{figure*}

We performed computational experiments comparing \AXE with prior baselines across four datasets, two models, seven explainers, and all eight prior evaluation metrics from table \ref{tab:prior_metrics}, adopting the standard OpenXAI benchmark \cite{agarwal2022openxai}, with results presented in figures \ref{fig:nn_openxai} and \ref{fig:lr_openxai}. Like OpenXAI, we use the German Credit \cite{german_credit}, COMPAS \cite{compas}, Adult Income \cite{adult_dataset}, and Home Equity Line of Credit (HELOC) \cite{heloc_dataset} datasets ($\mathcal{X}$), and run experiments with both linear regression and neural network models ($m$). We try the SHAP and LIME (perturbation based); SmoothGrad, Grad, Input \(\times\) Grad and Integrated Gradients (gradient based); and Random explainers ($\mathcal{E}$). We report results using the FA, RA, SA, SRA, RC, PRA, PGI, and PGU evaluation metrics ($q$).

We compare these baselines with $\AXE_n^k$. Instead of selecting a particular number of top-n features, we use the AUC trick described in section \ref{subsec:framework} for all evaluation metrics $q$. Instead of selecting a particular value for the \textit{k}-NN hyperparameter $k$, we report results for several values: $\text{AXE}^1$, $\text{AXE}^3$, $\text{AXE}^5$, and $\text{AXE}^9$. Each explainer type (perturbation, gradient, or random) is denoted with a different line-style. To compare evaluation metrics with each other, we standardize the final results for each explainer and evaluation metric using z-scores, because they may follow different scales. For instance, while ideal explanations for both \AXE and PGI have scores of 1.0, \AXE considers uninformative explanations to have values near 0.5, whereas PGI considers uninformative values to be near 0. Additionally, we also invert PGU values (denoted as (-)PGU) so that higher values are better, like the rest of our metrics.

For logistic regression models, we are able to use the ground-truth evaluation metrics because of the presence of model coefficients, which we adopt as ground-truth for every datapoint in the dataset, following previous benchmarks \cite{disagreement, agarwal2022openxai}. This approach is discussed in detail in section \ref{subsec:gt_failures}. For neural network models, we are only able to use sensitivity based metrics PGI and PGU, because of the lack of ground truth explanations. The results from the logistic regression comparisons can be seen in figure \ref{fig:lr_openxai} and from the neural network in figure \ref{fig:nn_openxai}.

From the plots in figures \ref{fig:nn_openxai} and \ref{fig:lr_openxai}, the \(\text{\AXE}^{1}\), \(\text{\AXE}^{3}\), \(\text{\AXE}^{5}\), and \(\text{\AXE}^{9}\) metrics can be seen to broadly agree with each other in score, further reinforcing the intuition from section \ref{subsec:demotivation} that \AXE is fairly robust to hyperparameter variations. The ground-truth oriented metrics (FA, SA, RA, SRA, RC, PRA) show significant disagreement with each other, as has been noted in the literature \cite{disagreement, disagreement_original, disagreement_new}.

Finally, as a simple check for evaluation metric validity, we focus on the behavior of the Random explainer. Ideally, a good evaluation framework would clearly and reliably distinguish this explainer from the others, however this does not seem to be the case for any previous evaluation frameworks. The sensitivity oriented metrics (PGI and PGU) rank the Random explainer particularly well. This is expected from prior work \cite{Riley2023-rs} and from our analysis from section \ref{subsec:demotivation} where we uncovered the dependence of PGI values on the neighborhood perturbation width hyperparameter. This further questions the use of feature sensitivity as an effective strategy to evaluate explanation quality, bolstering the importance of the \emph{on-manifold evaluation} principle.

\vspace{-0.3cm}
\section{Conclusion}
\label{sec:conclusion}
The ability to measure explanation quality is integral to developing better explanations, and to engender trust in existing XAI methods.

In section \ref{subsec:principles} we introduced three principles to guide the evaluation of explainers and local feature-importance model explanations: \textbf{local contextualization}, \textbf{model relativism}, and \textbf{on-manifold evaluation}, reflecting that explanations should be dependent on input datapoints, dependent on models, and independent of off-manifold behavior respectively. We constructed simple examples in sections \ref{subsec:gt_failures} and \ref{subsec:sens_failures} showcasing the violations of these principles by prior evaluation frameworks, and uncovering the absurdities of existing XAI evaluation frameworks -- such as comparing a single global explanation with local model explanations of different input datapoints. To operationalize the finding from human-centered user research that useful explanations are those that help users predict model behavior \cite{predictions_for_explanations}, in section \ref{subsec:framework} we proposed \AXE: a new ground-truth \textbf{Agnostic eXplanation Evaluation} framework, and in \ref{subsec:demotivation} we used a simple example to showcase \AXE in action and motivate the underlying design choice of \textit{k}-NN. Finally, in section \ref{subsec:adv_attack} we showed empirically how \AXE can be used to detect fairwashing of explanations -- to our knowledge the first evaluation metric to be able to do this perfectly, and in section \ref{subsec:openxai_experiments} we compared \AXE with prior baselines to show through computations that \AXE satisfies all three desirable principles of explanation evaluation.

This work has several implications for AI trustworthiness, fairness, and transparency. The lack of good selection processes to choose between explanations undermines trust not just in individual explanations and models, but in the field at large. It hinders practitioners from adopting XAI, and leads to unresolved problems about explanation disagreement in machine learning. We hope the principles introduced in this paper and the \AXE evaluation framework can help build a robust and stable foundation for local explanations in XAI.

\clearpage
\newpage

\begin{acks}

This work has been supported through research funding provided by the Wellcome Trust (grant no. 223765/Z/21/Z), Sloan Foundation (grant no. G-2021-16779), Department of Health and Social Care, EPSRC (grant no. EP/Y019393/1), and Luminate Group. Their funding supports the Trustworthiness Auditing for AI project and the Governance of Emerging Technologies research programme at the Oxford Internet Institute, University of Oxford. The donors had no role in the decision to publish or the preparation of this paper.

\end{acks}


\bibliographystyle{ACM-Reference-Format}
\bibliography{main}

\appendix



\end{document}